\pdfoutput=1
\documentclass[11pt]{article}

\usepackage{acl}

\usepackage{times}
\usepackage{latexsym}
\usepackage{dirtytalk}  % for quotes
\usepackage{listings}
\usepackage{amsmath}
\usepackage{array,multirow,graphicx}
\usepackage{float}
\usepackage{subcaption}

\usepackage[T1]{fontenc}
\usepackage[utf8]{inputenc}
\usepackage{microtype}
\usepackage{booktabs}
\usepackage{fancyvrb}
\usepackage{caption}
\usepackage{csquotes}

\usepackage{algorithm}
\usepackage[noend]{algpseudocode}

\usepackage{hyperref}

\newcommand{\pathpiece}{\textsc{PathPiece}}

\DeclareMathOperator{\rand}{rand}
\DeclareMathOperator{\append}{append}
\DeclareMathOperator{\len}{len}
\DeclareMathOperator{\ctc}{CTC}
\DeclareMathOperator{\reversed}{reversed}

\title{Tokenization Is More Than Compression}

\author{
Craig W. Schmidt\textsuperscript{\dag} \quad Varshini Reddy\textsuperscript{\dag} \quad Haoran Zhang\textsuperscript{\dag,\ddag} \quad Alec Alameddine\textsuperscript{\dag} \\
\smallskip \bf Omri Uzan\textsuperscript{\S} \quad Yuval Pinter\textsuperscript{\S} \quad Chris Tanner\textsuperscript{\dag,\P} \\
\begin{tabular}{cccc}
      \textsuperscript{\dag}Kensho Technologies & \textsuperscript{\ddag}Harvard Univ & \textsuperscript{\S}Ben-Gurion University &  \textsuperscript{\P}MIT \\
     Cambridge, MA  &  Cambridge, MA & Beer Sheva, Israel &  Cambridge, MA \\
\end{tabular} \\
\texttt{\small\{craig.schmidt,varshini.reddy,alec.alameddine,chris.tanner\}@kensho.com} \\ 
\texttt{\small haoran\_zhang@g.harvard.edu} \quad \texttt{\small\{omriuz@post,uvp@cs\}.bgu.ac.il} \\
}

\begin{document}

\maketitle
\begin{abstract}
Tokenization is a foundational step in natural language processing (NLP) tasks, bridging raw text and language models. Existing tokenization approaches like Byte-Pair Encoding (BPE) originate from the field of data compression, and it has been suggested that the effectiveness of BPE stems from its ability to condense text into a relatively small number of tokens. We test the hypothesis that fewer tokens lead to better downstream performance by introducing PathPiece, a new tokenizer that segments a document's text into the minimum number of tokens for a given vocabulary. Through extensive experimentation we find this hypothesis not to be the case, casting doubt on the understanding of the reasons for effective tokenization. To examine which other factors play a role, we evaluate design decisions across all three phases of tokenization: pre-tokenization, vocabulary construction, and segmentation, offering new insights into the design of effective tokenizers. Specifically, we illustrate the importance of pre-tokenization and the benefits of using BPE to initialize vocabulary construction. We train 64 language models with varying tokenization, ranging in size from 350M to 2.4B parameters, all of which are made publicly available.
\end{abstract}

\section{Introduction}
\label{sec:intro}

Tokenization is an essential step in NLP that translates human-readable text into a sequence of distinct tokens that can be subsequently used by statistical models~\cite{Grefenstette1999}. 
Recently, a growing number of studies have researched the effects of tokenization, both in an intrinsic manner and as it affects downstream model performance~\cite{singh-etal-2019-bert,bostrom-durrett-2020-byte,hofmann-etal-2021-superbizarre,hofmann-etal-2022-embarrassingly,limisiewicz-etal-2023-tokenization,zouhar-etal-2023-formal}.
To rigorously inspect the impact of tokenization, we consider tokenization as three distinct, sequential stages:
\begin{enumerate}
\item \textbf{Pre-tokenization:} an optional set of initial rules that restricts or enforces the creation of certain tokens (e.g., splitting a corpus on whitespace, thus preventing any tokens from containing whitespace).

\item \textbf{Vocabulary Construction:} the core algorithm that, given a text corpus $\mathcal{C}$ and desired vocabulary size $m$, constructs a vocabulary of tokens $t_k \in \mathcal{V}$, such that $|\mathcal{V}|=m$, while adhering to the pre-tokenization rules.

\item \textbf{Segmentation:} given a vocabulary $\mathcal{V}$ and a document $d$, segmentation determines how to split $d$ into a series of $K_d$ tokens $t_1,\dots,t_k,\dots,t_{K_d}$, with all $t_k \in \mathcal{V}$, such that the concatenation of the tokens strictly equals $d$. Given a corpus of documents $\mathcal{C}$, we will define the corpus token count (CTC) as the total number of tokens used in each segmentation, $\ctc(\mathcal{C}) = \sum_{d \in \mathcal{C}} K_d$.

As an example, segmentation might decide to split the text \texttt{intractable} into ``\texttt{int ract able}'', ``\texttt{in trac table}'', ``\texttt{in tractable}'', or ``\texttt{int~r act able}''.

We will refer to this step as segmentation, although in other works it is also called \say{inference} or even \say{tokenization}.
\end{enumerate}

The widely used Byte-Pair Encoding (BPE) tokenizer~\cite{sennrich-etal-2016-neural} originated in the field of data compression \cite{10.5555/177910.177914}. \citet{galle-2019-investigating} argues that it is effective because it compresses text to a short sequence of tokens. \citet{goldman2024unpacking} varied the number of documents in the tokenizer training data for BPE, and found a correlation between CTC and downstream performance. To investigate the hypothesis that having fewer tokens necessarily leads to better downstream performance, we design a novel tokenizer, \pathpiece{}, that, for a given document $d$ and vocabulary $\mathcal{V}$, finds a segmentation with the minimum possible $K_d$.  The \pathpiece{} vocabulary construction routine is a top-down procedure that heuristically minimizes CTC on a training corpus.  \pathpiece{} is ideal for studying the effect of CTC on downstream performance, as we can vary decisions at each tokenization stage.

We extend these experiments to the most commonly used tokenizers, focusing on how downstream task performance is impacted by the major stages of tokenization and vocabulary sizes. Toward this aim, we conducted experiments by training 64 language models (LMs): 54 LMs with 350M parameters; 6 LMs with 1.3B parameters; and 4 LMs with 2.4B parameters. We provide open-source, public access to \pathpiece{},\footnote{\url{https://github.com/kensho-technologies/pathpiece}} and our trained vocabularies and LMs.\footnote{\url{https://github.com/kensho-technologies/timtc_vocabs_models}}

\section{Preliminaries}
\label{sec:related_works}

\citet{ali2024tokenizer} and \citet{goldman2024unpacking} examined the effect of tokenization on downstream performance of LLM tasks, reaching opposite conclusions on the importance of CTC.
\citet{zouhar-etal-2023-tokenization} also find that low token count alone does not necessarily improve performance.
\citet{mielke2021words} give a survey of subword tokenization. 

\subsection{Pre-tokenization Methods} 
\label{subsec:pre_tok}

Pre-tokenization is a process of breaking text into chunks, which are then tokenized independently.
A token is not allowed to cross these pre-tokenization boundaries.
BPE, WordPiece, and Unigram all require new chunks to begin whenever a space is encountered.
If a space appears in a chunk, it must be the first character; hence, we will call this ``FirstSpace''.
Thus ``\texttt{\textvisiblespace{}New}'' is allowed but ``\texttt{New\textvisiblespace{}York}'' is not.
\citet{gow-smith-etal-2022-improving} examine treating spaces as individual tokens, which we will call ``Space'' pre-tokenization, while \citet{jacobs2022lost} suggest marking spaces at the end of tokens, and \citet{gow-smith-etal-2024-word} propose dispensing them altogether in some settings.
Llama~\cite{touvron2023llama} popularized the idea of having each digit always be an individual token, which we call ``Digit'' pre-tokenization.

\subsection{Vocabulary Construction} 
We focus on byte-level, lossless subword tokenization. Subword tokenization algorithms split text into word and subword units based on their frequency and co-occurrence patterns from their ``training'' data, effectively capturing morphological and semantic nuances in the tokenization process~\cite{Mikolov2011SUBWORDLM}.  

We analyze BPE, WordPiece, and Unigram as baseline subword tokenizers, using the implementations from HuggingFace\footnote{\url{https://github.com/huggingface/tokenizers}} with \texttt{ByteLevel} pre-tokenization enabled. We additionally study SaGe, a context-sensitive subword tokenizer, using version 2.0.\footnote{\url{https://github.com/MeLeLBGU/SaGe}}

\paragraph{Byte-Pair Encoding} ~\citet{sennrich-etal-2016-neural} introduced Byte-Pair Encoding (BPE), a bottom-up method where the vocabulary construction starts with single bytes as tokens.
It then merges the most commonly occurring pair of adjacent tokens in a training corpus into a single new token in the vocabulary.
This process repeats until the desired vocabulary size is reached.
Issues with BPE and analyses of its properties are discussed in \citet{bostrom-durrett-2020-byte,klein-tsarfaty-2020-getting,gutierrez-vasques-etal-2021-characters,yehezkel-pinter-2023-incorporating,saleva-lignos-2023-changes,liang-etal-2023-xlm, lian2024scaffoldbpeenhancingbytepair, chizhov2024bpegetspickyefficient, bauwens-delobelle-2024-bpe}.
\citet{zouhar-etal-2023-formal} build an \say{exact} algorithm which optimizes compression for BPE-constructed vocabularies.

\paragraph{WordPiece}

WordPiece is similar to BPE, except that it uses Pointwise Mutual Information (PMI)~\cite{bouma2009normalized} as the criteria to identify candidates to merge, rather than a count~\cite{wu2016googles, schuster-nakajima-wordpiece}.
PMI prioritizes merging pairs that occur together more frequently than expected, relative to the individual token frequencies.

\paragraph{Unigram Language Model}

Unigram works in a top-down manner, starting from a large initial vocabulary and progressively pruning groups of tokens that induce the minimum likelihood decrease of the corpus~\cite{kudo-2018-subword}.
This selects tokens to maximize the likelihood of the corpus, according to a simple unigram language model.

\paragraph{SaGe} \citet{yehezkel-pinter-2023-incorporating} proposed SaGe, a subword tokenization algorithm incorporating contextual information into an ablation loss via a skipgram objective.
SaGe also operates top-down, pruning from an initial vocabulary to a desired size.

\subsection{Segmentation Methods}
Given a tokenizer and a vocabulary of tokens, segmentation converts text into a series of tokens. We included all 256 single-byte tokens in the vocabulary of all our experiments, ensuring any text can be segmented without out-of-vocabulary issues.

Certain segmentation methods are tightly coupled to the vocabulary construction step, such as merge rules for BPE or the maximum likelihood approach for Unigram.
Others, such as the WordPiece approach of greedily taking the longest prefix token in the vocabulary at each point, can be applied to any vocabulary; indeed, there is no guarantee that a vocabulary will perform best downstream with the segmentation method used to train it~\cite{uzan-etal-2024-greed}.
Additional segmentation schemes include Dynamic Programming BPE~\cite{he-etal-2020-dynamic}, BPE-Dropout~\cite{provilkov-etal-2020-bpe}, and FLOTA~\cite{hofmann-etal-2022-embarrassingly}.

\section{\pathpiece{}}
\label{sec:pathpiece}

Several efforts over the last few years~\cite[][\textit{inter alia}]{galle-2019-investigating,zouhar-etal-2023-tokenization} have suggested that the empirical advantage of BPE as a tokenizer in many NLP applications, despite its unawareness of language structure, can be traced to its superior compression abilities, providing models with overall shorter sequences during learning and inference.
Inspired by this claim we introduce \pathpiece{}, a lossless subword tokenizer that, given a vocabulary $\mathcal{V}$ and document $d$, produces a segmentation minimizing the total number of tokens needed to split $d$.
We additionally provide a vocabulary construction procedure that, using this segmentation, attempts to find a $\mathcal{V}$ minimizing the corpus token count (CTC).\footnote{An extended description is given in \autoref{sec:full_pathpiece}.}
\pathpiece{} provides an ideal testing laboratory for the compression hypothesis by virtue of its maximally efficient segmentation.

\subsection{Segmentation}

\pathpiece{} requires that all single-byte tokens are included in vocabulary $\mathcal{V}$ to run correctly.
\pathpiece{} works by finding a shortest path through a directed acyclic graph (DAG), where each byte $i$ of training data forms a node in the graph,
and two nodes $j$ and $i$ contain a directed edge if the byte segment $[j,i]$ is a token in $\mathcal{V}$.
% and there is an edge if the $w$ byte sequence ending at $i$ is a token in $\mathcal{V}$.  
We describe \pathpiece{} segmentation in Algorithm~\ref{alg:pathpiece3}, where $L$ is a limit on the maximum width of a token in bytes, which we set to 16.
It has a complexity of $O(nL)$, which follows directly from the two nested \texttt{for}-loops.
For each byte $i$ in $d$, it computes the shortest path length $pl[i]$ in tokens up to and including byte $i$, and the width $wid[i]$ of a token with that shortest path length.
In choosing $wid[i]$, ties between multiple tokens with the same shortest path length $pl[i]$ can be broken randomly, or the one with the longest $wid[i]$ can be chosen, as shown here.\footnote{Random tie-breaking, which can be viewed as a form of subword regularization, is presented in \autoref{sec:full_pathpiece}. Some motivation for selecting the longest token is due to the success of FLOTA~\cite{hofmann-etal-2022-embarrassingly}.}
Then, a backward pass constructs the shortest possible segmentation from the $wid[i]$ values computed in the forward pass.

\begin{algorithm}[hbt!]
\begin{algorithmic}[1]
\small
\Procedure {PathPiece}{$d, \mathcal{V}, L$}
\State $n \leftarrow \len(d)$ 
\Comment{document length}

\State $pl[1:n] \leftarrow \infty$
\Comment{shortest path length}
\State $wid[1:n] \leftarrow 0$
\Comment{shortest path tok width}
    
\For {$e \leftarrow 1, n$}
\Comment{token end}
    \For {$w \leftarrow 1,L$}
        \Comment{token width}
        \State $s \leftarrow e - w + 1$
        \Comment{token start}
    
        \If {$s \ge 1$}
        \Comment{$s$ in range}
            \If {$d[s:e] \in \mathcal{V}$}
                \If {$s = 1$}
                \Comment{1 tok path}
                    \State $pl[e] \leftarrow 1$
                    \State $wid[e] \leftarrow w$
                \Else
                    \State $nl \leftarrow pl[s-1] + 1$ 
                    \If {$nl \le pl[e]$}
                            \State $pl[e] \leftarrow nl$
                            \State $wid[e] \leftarrow w$
                    \EndIf
                \EndIf
            \EndIf
        \EndIf    
    \EndFor
\EndFor
\State $T \leftarrow [\,]$
\Comment{output token list}
\State $e \leftarrow n$
\Comment{start at end of $d$}
\While {$e \ge 1$}
    \State $s \leftarrow e - wid[e] + 1$
    \Comment{token start}    
    \State $T.\append(d[s:e])$
    \Comment{append token}
    \State $e \leftarrow  e - wid[e]$
    \Comment{back up a token}
\EndWhile
\State \textbf{return} {$\reversed(T)$}
\Comment{reverse order}
\EndProcedure
\end{algorithmic}
\caption{\pathpiece{} segmentation.}\label{alg:pathpiece3}
\end{algorithm}

\subsection{Vocabulary Construction}

 \pathpiece{}'s vocabulary is built in a top-down manner, attempting to minimize the corpus token count (CTC), by starting from a large initial vocabulary $\mathcal{V}_{0}$ and iteratively omitting batches of tokens.
The $\mathcal{V}_{0}$ may be initialized from the most frequently occurring byte $n$-grams in the corpus, or from a large vocabulary trained by BPE or Unigram.
We enforce that all single-byte tokens remain in the vocabulary and that all tokens are $L$ bytes or shorter.

For a \pathpiece{} segmentation $t_1,\dots,t_{K_d}$ of a document $d$ in the training corpus $\mathcal{C}$, we would like to know the increase in the overall length of the segmentation $K_d$ after omitting each token $t$ from our vocabulary %$\mathcal{V}\setminus \{t\}$
and then recomputing the segmentation.
Tokens with a low overall increase are good candidates to remove from the vocabulary.

To avoid the very expensive $O(nL|\mathcal{V}|)$ computation of each segmentation from scratch, we make a simplifying assumption that allows us to compute these increases more efficiently:
we omit a specific token $t_k$, for $k \in [1,\dots,K_d]$ in the segmentation of a particular document $d$, and compute the minimum increase $MI_{kd} \ge 0$ in the total tokens $K_d$ from not having that token $t_k$ in the segmentation of $d$.
We then aggregate these token count increases $MI_{kd}$ for each token $t \in \mathcal{V}$.
We can compute the $MI_{kd}$ without actually re-segmenting any documents, by reusing the shortest path information computed by Algorithm~\ref{alg:pathpiece3} during segmentation. 

Any segmentation not containing $t_k$ must either contain a token boundary somewhere inside of $t_k$ breaking it in two, or it must contain a token that entirely contains $t_k$ as a superset. 
We enumerate all occurrences for these two cases, and we find the minimum increase $MI_{kd}$ among them.
Let $t_k$ start at index $s$ and end at index $e$, inclusive.
Path length $pl[j]$ represents the number of tokens required for the shortest path up to and including byte $j$.
We also run Algorithm~\ref{alg:pathpiece3} backwards on $d$, computing a similar vector of backwards path lengths $bpl[j]$, representing the number of tokens on a path from the end of the data up to and including byte $j$.
The minimum length of a segmentation with a token boundary after byte $j$ is thus:
\begin{equation}
    K_j^{b} = pl[j] + bpl[j+1].
\end{equation}
We have added an extra constraint on the shortest path, that there is a break at $j$, so clearly $K_j^{b} \ge K_d$. The minimum increase for the case of having a token boundary within $t_k$ is thus:
\begin{equation}
    MI_{kd}^{b} = \min_{j=s,\dots,e-1} {K_j^{b} - K_d}.
\end{equation}

The minimum increase from omitting $t_k$ could also be from a segmentation containing a strict superset of $t_k$.  Let this superset token be $t_k'$, with start $s'$ and end $e'$ inclusive.  To be a strict superset entirely containing $t_k$, then either $s' < s$ and $e' \ge e$, or $s' \le s$ and $e' > e$, subject to the constraint that the width $w' = e' - s' + 1 \le L$.  In this case, the minimum length when using the superset token $t_k'$ would be:
\begin{equation}
    K_{t_k'}^{s} = pl[s'-1] + bpl[e'+1] + 1,
\end{equation}
which is the path length to get to the byte before $t_k'$, plus the path length from the end of the data backwards to the byte after $t_k'$, plus 1 for the token $t_k'$ itself.

We retain a list of the widths of the tokens ending at each byte.\footnote{See the expanded explanation in \autoref{sec:full_pathpiece} for details.} The set of superset tokens $S$ can be found by examining the potential $e'$, and then seeing if the tokens ending at $e'$ form a strict superset.  Similar to the previous case, we can compute the minimum increase from replacing $t_k$ with a superset token by taking the minimum increase over the superset tokens $S$:
\begin{equation}
    MI_{kd}^{s} = \min_{t_k' \in S} {K_{t_k'}^{s}  - K_d}.
\end{equation}

We then aggregate over the documents to get the overall increase for each $t \in \mathcal{V}$:
%\vspace{-5pt}
\begin{equation}
    MI_t = \sum_{d \in \mathcal{C}} \sum_{k=1 | t_k = t}^{K_d} \min(MI_{kd}^{b},MI_{kd}^{s}).
\end{equation}
%\vspace{-5pt}
 
One iteration of this vocabulary construction procedure will have complexity $O(n L^2)$.\footnotemark[\value{footnote}]

\subsection{Connecting \pathpiece{ and Unigram}}
We note a connection between \pathpiece{} and Unigram.
In Unigram, the probability of a segmentation $t_1,\dots,t_{K_d}$ is the product of the unigram token probabilities $p(t_k)$:

\begin{equation}
    P(t_1,\dots,t_{K_d}) = \prod_{k=1}^{K_d} p(t_k).
\end{equation}

Taking the negative $\log$ of this product converts the objective from maximizing the likelihood to minimizing the sum of $-\log(p(t_k))$ terms.
While Unigram is solved by the \citet{Viterbi1967} algorithm, it can also be solved by a weighted version of \pathpiece{} with weights of $-\log(p(t_k))$.
Conversely, a solution minimizing the number of tokens can be found in Unigram by taking all $p(t_k) := 1/|\mathcal{V}|$.

\section{Experiments}
\label{sec:exp}

We used the Pile corpus~\cite{gao2020pile, DBLP:journals/corr/abs-2201-07311} for language model pre-training, which contains 825GB of English text data from 22 high-quality datasets.
We constructed the tokenizer vocabularies over the MiniPile dataset~\cite{kaddour2023minipile}, a 6GB subset of the Pile.
We use the MosaicML Pretrained Transformers (MPT) decoder-only language model architecture.\footnote{\url{https://github.com/mosaicml/llm-foundry}}
\autoref{section:lang_model_params} gives the full set of model parameters, and \autoref{sec:model_convergence} discusses model convergence.

\subsection{Downstream Evaluation Tasks}
\label{subsec:benchmark_tasks}

To evaluate and analyze the performance of our tokenization process, we select 10 benchmarks from \texttt{lm-evaluation-harness}~\cite{eval-harness}.\footnote{\url{https://github.com/EleutherAI/lm-evaluation-harness}}
These are all multiple-choice tasks with 2, 4, or 5 options, and were run with 5-shot prompting.
We use arc\_easy~\cite{arc_easy}, copa~\cite{copa}, hendrycksTests-marketing~\cite{hendrycksTests}, hendrycksTests-sociology~\cite{hendrycksTests}, mathqa~\cite{mathqa}, piqa~\cite{piqa}, qa4mre\_2013~\cite{qa4mre}, race~\cite{race}, sciq~\cite{sciq}, and wsc273~\cite{winograd}.
\autoref{section:descr_benchmark_tasks} gives a full description of these tasks.

\subsection{Tokenization Stage Variants}
\label{subsec:options}

We conduct the 18 experimental variants listed in \autoref{tab:overallavg_grouped}, each repeated at the vocabulary sizes of 32,768, 40,960, and 49,152.\footnote{These sizes were selected because vocabularies in the 30k to 50k range are the most common amongst language models within the HuggingFace Transformers library, \url{https://huggingface.co/docs/transformers/}. \citet{ali2024tokenizer} recently examined the effect of vocabulary sizes and found 33k and 50k sizes performed better on English language tasks than larger sizes.}
For baseline vocabulary creation methods, we used BPE, Unigram, WordPiece, and SaGe.
We also consider two variants of \pathpiece{} where ties in the shortest path are broken either by the longest token (\textsc{PathPieceL}), or randomly (\textsc{PathPieceR}). For the vocabulary initialization required by \pathpiece{} and SaGe, we experimented with the most common $n$-grams, as well as with a large initial vocabulary trained with BPE or Unigram. We also varied the pre-tokenization schemes for \pathpiece{} and SaGe, using either no pre-tokenization or combinations of \say{FirstSpace}, \say{Space}, and \say{Digit} described in \S\ref{subsec:pre_tok}. Tokenizers usually use the same segmentation approach used in vocabulary construction.
\textsc{PathPieceL}'s shortest path segmentation can be used with any vocabulary, so we apply it to vocabularies trained by BPE and Unigram.
We also apply a Greedy left-to-right longest-token segmentation approach to these vocabularies.

\section{Results}
\label{sec:results}
\autoref{tab:overallavg_grouped} reports the downstream performance across all our experimental settings.\footnote{The same table sorted by rank is in \autoref{tab:overallavg} of \autoref{app:experimental_results}.
The comprehensive results for the ten downstream tasks, for each of the 350M parameter models, are given in \autoref{app:experimental_results}.} A random baseline for these 10 tasks yields 32\%.
The \textsc{Overall Avg} column indicates the average results over the three vocabulary sizes.
The \textsc{Rank} column refers to the rank of each variant with respect to the \textsc{Overall Avg} column (Rank 1 is best), which we will sometimes use as a succinct way to refer to a variant.

\begin{table*}[!htb]
    \centering
    \small
    \begin{tabular}{cllllrrrr}
    \toprule
\textbf{Rank}	&	\textbf{Vocab Constr}	&	\textbf{Init Voc}	&	\textbf{Pre-tok}	&	\textbf{Segment}	&	\textbf{Overall}	&	\textbf{32,768}	&	\textbf{40,960}	&	\textbf{49,152}	\\

\midrule

1	&	\multirow{4}{*}{PathPieceL}	&	BPE	&	FirstSpace	&	\multirow{4}{*}{PathPieceL}	&	\textbf{49.4}	&	\textbf{49.3}	&	49.4	&	49.4	\\
9	&	&	Unigram	& FirstSpace	&	&	48.0	&	47.0	&	48.5	&	48.4	\\
15	&	&	$n$-gram	&	FirstSpDigit	&	&	44.8	&	44.6	&	44.9	&	45.0	\\
16	&	&	$n$-gram	&	FirstSpace	&	&	44.7	&	44.8	&	45.5	&	43.9	\\
\midrule
2	&	\multirow{3}{*}{Unigram}	&		&	\multirow{3}{*}{FirstSpace}	&	Likelihood	&	49.0	&	49.2	&	49.1	&	48.8	\\
7	&	&		&	&	Greedy	&	48.3	&	47.9	&	48.5	&	48.6	\\
17	&	&		&	&	PathPieceL	&	43.6	&	43.6	&	43.1	&	44.0	\\
\midrule
3	&	\multirow{3}{*}{BPE}	&		&	\multirow{3}{*}{FirstSpace}	&	Merge	&	49.0	&	49.0	&	\textbf{50.0}	&	48.1	\\
4	&	&		&	&	Greedy	&	49.0	&	48.3	&	49.1	&	\textbf{49.5}	\\
13	&	&		&	&	PathPieceL	&	46.5	&	45.6	&	46.7	&	47.2	\\
\midrule
5	&	WordPiece	&		&	FirstSpace	&	Greedy	&	48.8	&	48.5	&	49.1	&	48.8	\\
\midrule
6	&	\multirow{4}{*}{SaGe}	&	BPE	& FirstSpace	&	\multirow{4}{*}{Greedy}	&	48.6	&	48.0	&	49.2	&	48.8	\\
% 7	&	Unigram	&		&	FirstSpace	&	Greedy	&	48.33	&	47.87	&	48.50	&	48.64	\\
8	&	&	$n$-gram	&	FirstSpace &	&	48.0	&	47.5	&	48.5	&	48.0	\\
% 9	&	PathPieceL	&	Unigram	&	FirstSpace	&	PathPieceL	&	47.96	&	46.92	&	48.51	&	48.44	\\
10	&	&	Unigram	& FirstSpace	&	&	47.7	&	48.4	&	46.9	&	47.8	\\
11	&	&	$n$-gram	&	FirstSpDigit	&	&	47.5	&	48.4	&	46.9	&	47.2	\\
\midrule
12	&	\multirow{3}{*}{PathPieceR}	&	\multirow{3}{*}{$n$-gram}	&	SpaceDigit	&	\multirow{3}{*}{PathPieceR}	&	46.7	&	47.5	&	45.4	&	47.3	\\
% 13	&	BPE	&		&	FirstSpace	&	PathPieceL	&	46.49	&	45.56	&	46.66	&	47.23	\\
14	&	&	&	FirstSpDigit	&	&	45.5	&	45.3	&	45.8	&	45.5	\\
% 15	&	PathPieceL	&	$n$-gram	&	FirstSpDigit	&	PathPieceL	&	44.82	&	44.59	&	44.86	&	45.00	\\
% 16	&	PathPieceL	&	$n$-gram	&	FirstSpace	&	PathPieceL	&	44.74	&	44.78	&	45.51	&	43.92	\\
% 17	&	Unigram	&		&	FirstSpace	&	PathPieceL	&	43.56	&	43.63	&	43.07	&	43.97	\\
18	&	&	&	None	&	&	43.2	&	43.5	&	44.0	&	42.2	\\

\midrule
		&	Random	&		&		&		&	32.0 & 32.0 & 32.0 & 32.0	\\

    \bottomrule
    \end{tabular}
       \caption{Summary of 350M parameter model downstream accuracy (\%) across 10 tasks. The \protect\say{Overall} column averages across the three vocabulary sizes. The \protect\say{Rank} column refers to the Overall column, best to worst.}
    \label{tab:overallavg_grouped}
    
\end{table*}

\subsection{Vocabulary Size}
\label{subsec:vocabsize}

\begin{figure}[ht]
    \centering
    \includegraphics[width=\columnwidth]{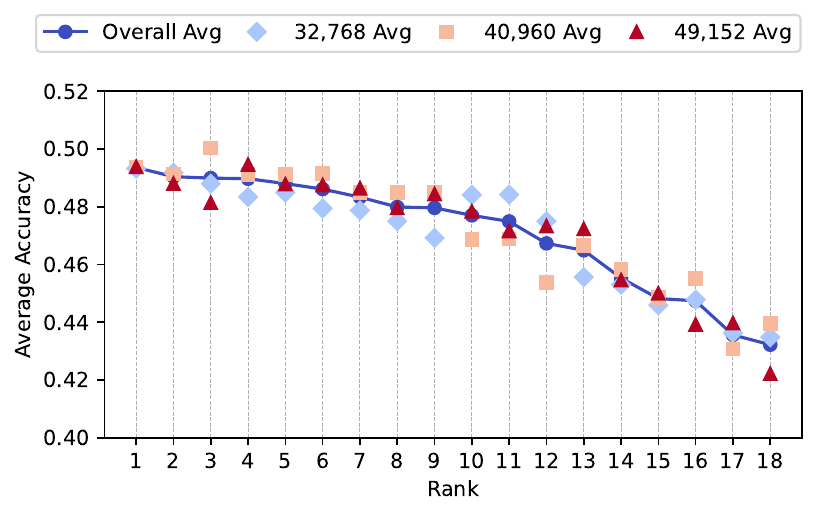}
    \caption{Effect of vocabulary size on downstream performance. For each tokenizer variant, we show the overall average, along with the three averages by vocabulary size, labeled according to the ranks in \autoref{tab:overallavg_grouped}.}
    \label{fig:effect_of_vocab_size}
\end{figure}
\autoref{fig:effect_of_vocab_size} gives the overall average, along with the individual averages, for each of the three vocabulary sizes for each variant, labeled according to the rank from \autoref{tab:overallavg_grouped}. We observe that there is a high correlation between downstream performance at different vocabulary sizes.
The pairwise $R^2$ values for the accuracy of the 32,768 and 40,960 runs was 0.750; between 40,960 and 49,152 it was 0.801; and between 32,768 and 49,152 it was 0.834. This corroborates the effect shown graphically in \autoref{fig:effect_of_vocab_size} that vocabulary size is not a crucial decision over this range of sizes. Given this high degree of correlation, we focus our analysis on the overall average accuracy.  This averaging removes some of the variance amongst individual language model runs. Thus, unless specified otherwise, our analyses present performance averaged over vocabulary sizes.

\subsection{Overall performance}
\label{subsec:overall}

To determine which of the differences in the overall average accuracy in \autoref{tab:overallavg_grouped} are statistically significant, we conduct a one-sided Wilcoxon signed-rank test~\cite{wilcoxon1945individual} on the paired differences of the 30 accuracy scores (three vocabulary sizes over ten tasks), for each pair of variants.
All $p$-values reported in this paper use this test.

\begin{figure}[ht]
    \centering
    \includegraphics[width=\columnwidth]{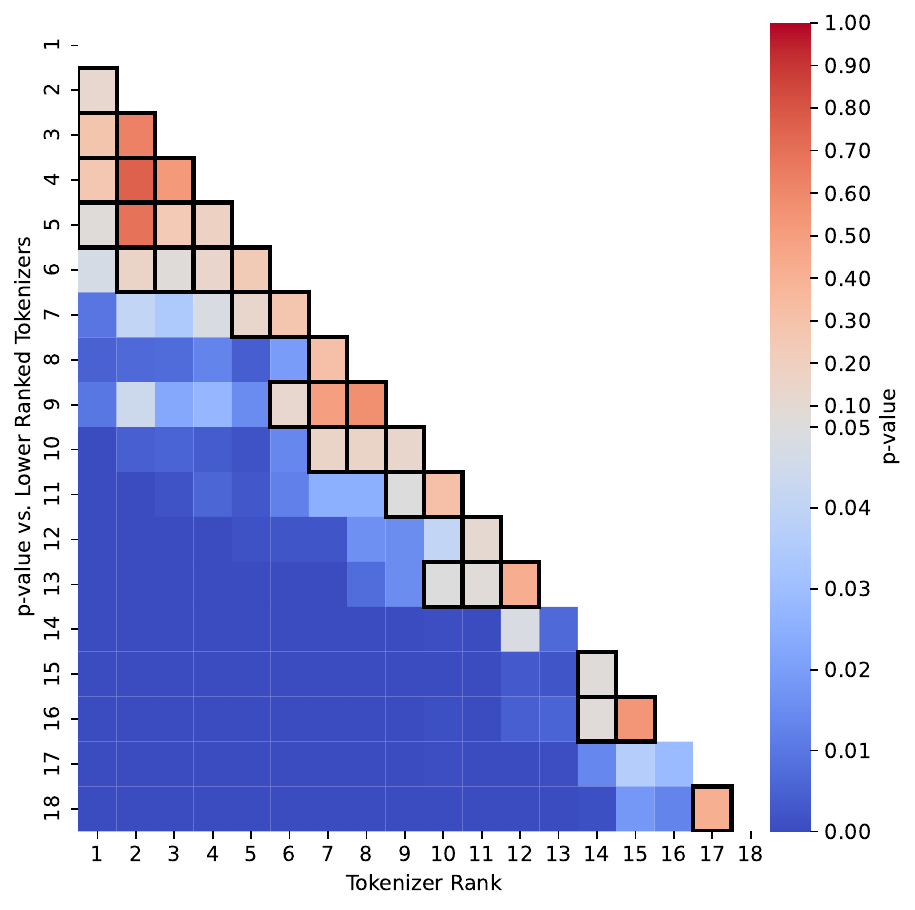}
    \caption{Pairwise $p$-values for 350M model results. Boxes outlined in black represent $p$ > 0.05. The top 6 tokenizers are all competitive, and there is no statistically significantly best approach. }
    \label{fig:overall_p_values}
\end{figure}

\autoref{fig:overall_p_values} displays all pairwise $p$-values in a color map.
Each column designates a tokenization variant by its rank in \autoref{tab:overallavg_grouped}, compared to all the ranks below it.
A box is outlined in black if $p > 0.05$, where we cannot reject the null.
While \textsc{PathPieceL}-BPE had the highest overall average on these tasks, the top five tokenizers, \textsc{PathPieceL}-BPE, Unigram, BPE, BPE-Greedy, and WordPiece do not have any other row in \autoref{fig:overall_p_values} significantly different from them.
Additionally, SaGe-BPE (rank 6) is only barely worse than \textsc{PathPieceL}-BPE ($p$~=~0.047), and should probably be included in the list of competitive tokenizers.
Thus, our first key result is that there is no tokenizer algorithm better than all others to a statistically significant degree.

All the results reported thus far are for language models with identical architectures and 350M parameters.
To examine the dependency on model size, we trained larger models of 1.3B parameters for six of our experiments, and 2.4B parameters for four of them.
In the interest of computational time, these larger models were only trained with a single vocabulary size of 40,960. In \autoref{fig:larger_size} in \autoref{subsec:modelsize}, we report models' average performance across 10 tasks.
See \autoref{fig:checkpoint_graph} in \autoref{sec:model_convergence} for an example checkpoint graph at each model size.
The main result from these models is that the relative performance of the tokenizers does vary by model size, and that there is a group of high performing tokenizers that yield comparable results.
This aligns with our finding that the top six tokenizers are not statistically better than one another at the 350M model size.

\subsection{Corpus Token Count vs Accuracy}\label{subsec:versus}

\begin{figure}
    \centering
    \includegraphics[width=\columnwidth]{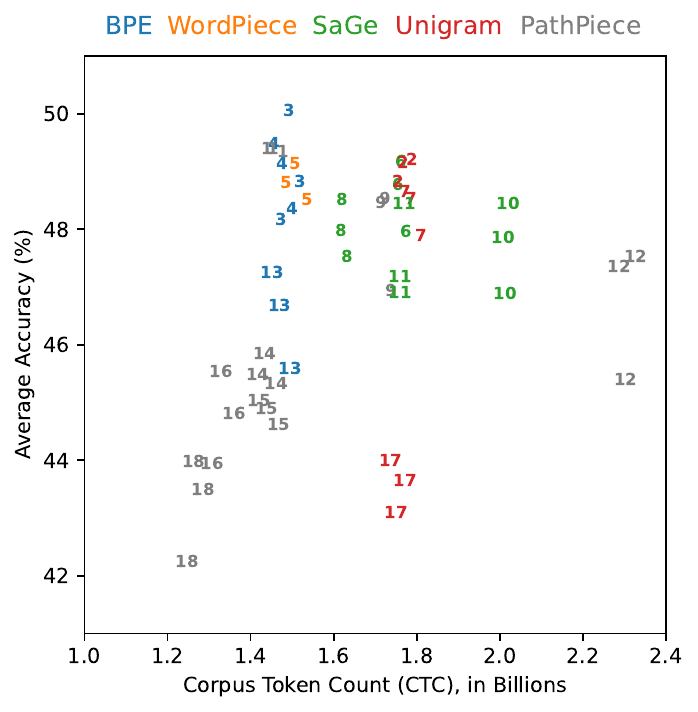}
    \caption{Effect of corpus token count (CTC) vs average accuracy of individual vocabulary sizes.}
    \label{fig:total_tok_vs_accuracy}
\end{figure}

\autoref{fig:total_tok_vs_accuracy} shows the corpus token count (CTC) versus the accuracy of each vocabulary size, given in \autoref{tab:fig3data}. We do not find a straightforward relationship between the two. \citet{ali2024tokenizer} recently examined the relationship between CTC and downstream performance for three different tokenizers, and also found it was not correlated on English language tasks.

The two models with the highest CTC are \pathpiece{} with Space pre-tokenization (12), which is to be expected given each space is its own token, and SaGe with an initial Unigram vocabulary (10). The Huggingface Unigram models in \autoref{fig:total_tok_vs_accuracy} had significantly higher CTC than the corresponding BPE models, unlike \citet{bostrom-durrett-2020-byte} and \citet{gow-smith-etal-2022-improving}, which report a difference of only a few percent with SentencePiece Unigram. \citet{ali2024tokenizer} point out that due to differences in pre-processing, the Huggingface Unigram tokenizer behaves quite differently than the SentencePiece Unigram tokenizer, which may explain this discrepancy.

In terms of accuracy, \pathpiece{} with no pre-tokenization (18) and Unigram with \pathpiece{} segmentation (17) both did quite poorly. Notably, the range of CTC is quite narrow within each vocabulary construction method, even while changes in pre-tokenization and segmentation lead to significant accuracy differences. While there are confounding factors present in this chart (e.g., pre-tokenization, vocabulary initialization, and that more tokens allow for additional computations by the downstream model) it is difficult to discern any trend that lower CTC leads to improved performance. If anything, there seems to be an inverted U-shaped curve with respect to the CTC and downstream performance. The Pearson correlation coefficient between CTC and average accuracy was found to be 0.241. Given that a lower CTC value signifies greater compression, this result suggests a weak negative relationship between the amount of compression and average accuracy.

\citet{zouhar-etal-2023-tokenization} introduced an information-theoretic measure based on R\'enyi efficiency that correlates with downstream performance for their application.\footnote{Except, so far, for a family of adversarially-created tokenizers~\cite{cognetta-etal-2024-two}.}
It has an order parameter $\alpha$, with a recommended value of 2.5.
We present the R\'enyi efficiencies and CTC for all models in \autoref{tab:fig3data} in \autoref{app:experimental_results}, and summarize their Pearson correlation with average accuracy in \autoref{tab:correlations}.
For the data of \autoref{fig:total_tok_vs_accuracy}, all the correlations for various $\alpha$ also have a weak negative association.
They are slightly less negative than the association for CTC, although it is not nearly as large as the benefit they saw over sequence length in their application.
We note the strong relationship between compression and R\'enyi efficiency, as the Pearson correlation of CTC and R\'enyi efficiency with $\alpha$=2.5 is $-$0.891.

\begin{table}
    \centering
    \small
    \begin{tabular}{lr}
    \toprule
\textbf{Comparison} & \textbf{Pearson Correlation} \\
\midrule
CTC and Ave Acc&  0.241 \\
\midrule
R\'enyi Eff and Ave Acc ($\alpha$=1.5)   &  $-$0.221   \\
R\'enyi Eff and Ave Acc ($\alpha$=2.0)   &  $-$0.169 \\
R\'enyi Eff and Ave Acc ($\alpha$=2.5)   & $-$0.151 \\
R\'enyi Eff and Ave Acc ($\alpha$=3.0)    &   $-$0.144 \\
R\'enyi Eff and Ave Acc ($\alpha$=3.5)   & $-$0.141 \\
\midrule
CTC and R\'enyi Eff ($\alpha$=2.5) & $-$0.891 \\
   \bottomrule
    \end{tabular}
   \caption{Pearson Correlation of CTC and Average Accuracy, or R\'enyi efficiency for various orders $\alpha$ with Average Accuracy, or CTC and R\'enyi efficiency at $\alpha=2.5$.}
    \label{tab:correlations}
    
\end{table}

By varying aspects of BPE, \citet{galle-2019-investigating} and \citet{goldman2024unpacking} suggests we should expect downstream performance to decrease with CTC, while in contrast \citet{ali2024tokenizer} did not find a strong relation when varying the tokenizer. Our extensive results varying a number of stages of tokenization suggest it is not \textit{inherently} beneficial to use fewer tokens. Rather, the particular way that the CTC is varied can lead to different conclusions.

\section{Analysis}
\label{sec:analysis}

We now analyze the results across the various experiments in a more controlled manner. Our experiments allow us to examine changes in each stage of tokenization, holding the rest constant, revealing design decisions making a significant difference.\footnote{Appendix \ref{app:additional_analysis} contains additional analysis}

\subsection{Pre-tokenization}
\label{subsec:space_pre_tokenization}

For \textsc{PathPieceR} with an $n$-gram initial vocabulary, we can isolate pre-tokenization. \pathpiece{} is efficient enough to process entire documents with no pre-tokenization, giving it full freedom to minimize the corpus token count (CTC).

\begin{table*}[ht]
    \centering
    \small
    \begin{tabular}{rll}
    \toprule
\textbf{Rank}	&	\textbf{Pre-tokenization} & \textbf{Example} \\

% \midrule
% 12  &   SpaceDigit  & \verb!The| |valuation| |is| |estimated| |to| |be| |$|2|1|3|M! \\
% 14 &    FirstSpDigit    & \verb!The| valuation| is| estimated| to| be| $|2|1|3|M! \\
% 18 & None   & \verb!The| valu|ation is| estimated| to b|e $|2|1|3|M! \\
% \midrule
% 2 & Unigram  & \verb!The| |valuation| is| estimated| to| be| |$|213|M! \\
% 3 & BPE &  \verb!The| valuation| is| estimated| to| be| $|213|M! \\

\midrule
12  &   SpaceDigit  & \texttt{The \textvisiblespace{} valuation \textvisiblespace{} is \textvisiblespace{} estimated \textvisiblespace{} to \textvisiblespace{} be \textvisiblespace{} \$ 2 1 3 M} \\
14 &    FirstSpDigit    & \texttt{The \textvisiblespace{}valuation \textvisiblespace{}is \textvisiblespace{}estimated \textvisiblespace{}to \textvisiblespace{}be \textvisiblespace{}\$ 2 1 3 M} \\
18 & None   & \texttt{The \textvisiblespace{}valu ation\textvisiblespace{}is \textvisiblespace{}estimated \textvisiblespace{}to\textvisiblespace{}b e\textvisiblespace{}\$ 2 1 3 M} \\
% \midrule
% 2 & Unigram  & \texttt{The \textvisiblespace{} valuation \textvisiblespace{}is \textvisiblespace{}estimated \textvisiblespace{}to \textvisiblespace{}be \textvisiblespace{} \$ 213 M} \\
% 3 & BPE &  \texttt{The \textvisiblespace{}valuation \textvisiblespace{}is \textvisiblespace{}estimated \textvisiblespace{}to \textvisiblespace{}be \textvisiblespace{}\$ 213 M} \\

    \bottomrule
    \end{tabular}
   \caption{Example \pathpiece{} tokenizations of ``The valuation is estimated to be \$213M''; vocabulary size of 32,768.}
    \label{tab:example_tok}
    
\end{table*}

\begin{figure}%[ht]
    \centering
    \includegraphics[width=\columnwidth]{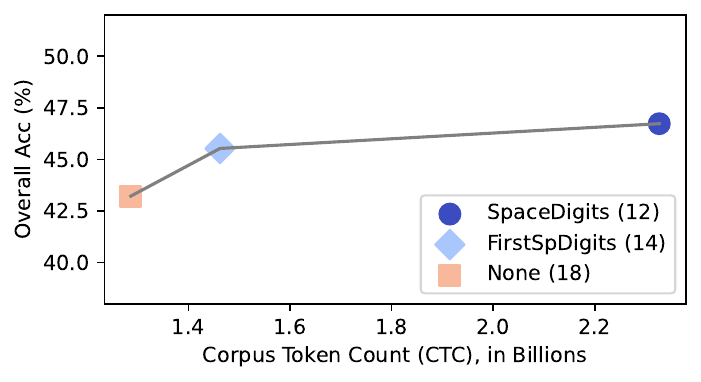}
    \caption{The impact of pre-tokenization on Corpus Token Count (CTC) and Overall Accuracy.
    Ranks in parentheses refer to performance in \autoref{tab:overallavg_grouped}.}
    \label{fig:total_tokens_vs_acc}
\end{figure}

Adding pre-tokenization constrains \pathpiece{}'s ability to minimize tokens, giving a natural way to vary the number of tokens. \autoref{fig:total_tokens_vs_acc} shows that \pathpiece{} minimizes the number of tokens used over a corpus when trained with no pre-tokenization (18). The other variants restrict spaces to either be the first character of a token (14), or their own token (12).\footnote{These two runs also used Digit pre-tokenization where each digit is its own token.} Consider the example \pathpiece{} tokenization in \autoref{tab:example_tok} for the three pre-tokenization methods.
The \textsc{None} mode uses the word-boundary-spanning tokens ``\texttt{ation\textvisiblespace{}is}'', ``\texttt{\textvisiblespace{}to\textvisiblespace{}b}'', and ``\texttt{e\textvisiblespace{}\$}''.
The lack of morphological alignment demonstrated in this example is likely more important to downstream model performance than a simple token count.

In \autoref{fig:total_tokens_vs_acc} we observe a statistically significant increase in overall accuracy for our downstream tasks, as a function of CTC. \citet{gow-smith-etal-2022-improving} found that Space pre-tokenization lead to worse performance, while removing the spaces entirely helps\footnote{Although omitting the spaces entirely does not lead to a reversible tokenization as we have been considering.}. Thus, this particular result may be specific to \textsc{PathPieceR}.

% TODO: if we don't include these they should probably move to the appendix
%\\begin{figure}[ht]
%     \centering
%     \includegraphics[width=\columnwidth]{graphics/pieces_space_pretok.pdf}
%     \captionsetup{justification=centering}
%     \caption{Pre-tokenization from PathPieceR, $n$-gram.\\
%     Pairwise $p$-values between the pairs of runs are \\
%     $p$(12,14)=0.048, $p$(12,18)=3.1e-5, $p$(14,18)=9.1e-4.} 
%     \label{fig:pieces_space_pretok}
% \end{figure}

\subsection{Vocabulary Construction}

\begin{figure}
    \centering
    \includegraphics[width=6cm]{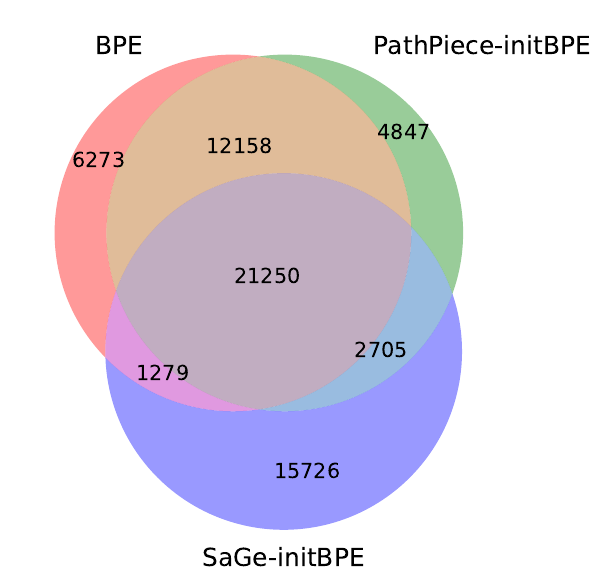}
    \caption{Venn diagram comparing 40,960 token vocabularies of BPE, PathPieceL and SaGe -- the latter two were both initialized from a BPE vocabulary of 262,144.}
    \label{fig:venn_diagram_bpe}
\end{figure}

One way to examine the effects of vocabulary construction is to compare the resulting vocabularies of top-down methods trained using an initial vocabulary to the method itself.
\autoref{fig:venn_diagram_bpe} presents an area-proportional Venn diagram of the overlap in 40,960-sized vocabularies between BPE (6) and variants of \textsc{PathPieceL} (1) and SaGe (6) that were trained using an initial BPE vocabulary of size $2^{18} = 262,144$.\footnote{See \autoref{fig:venn_diagram_unigram} in Appendix~\ref{sub:venn_unigram} for analogous results for Unigram, which behaves similarly.}
While BPE and \textsc{PathPieceL} overlap considerably, SaGe produces a more distinct set of tokens.

\subsection{Initial Vocabulary}
\label{subsec:initvocab}

\pathpiece{}, SaGe, and Unigram all require an initial vocabulary.\footnote{The HuggingFace Unigram implementation starts with the one millionp $n$-grams, but sorted according to the count times the length of the token, introducing a bias toward longer tokens.}
For \pathpiece{} and SaGe, we experimented with initial vocabularies of size 262,144 constructed from either the most frequent $n$-grams, or trained using either BPE or Unigram. For \textsc{PathPieceL}, using a BPE initial vocabulary (1) is statistically better than both Unigram (9) and $n$-grams (16), with $p \le 0.01$. Using an $n$-gram initial vocabulary leads to the lowest performance, with statistical significance.
Comparing ranks 6, 8, and 10 reveals the same pattern for SaGe, although the difference between 8 and 10 is not significant.

\subsection{Effect of Model Size}
\label{subsec:modelsize}

\begin{figure}[ht]
    \centering
    \includegraphics[width=\columnwidth]{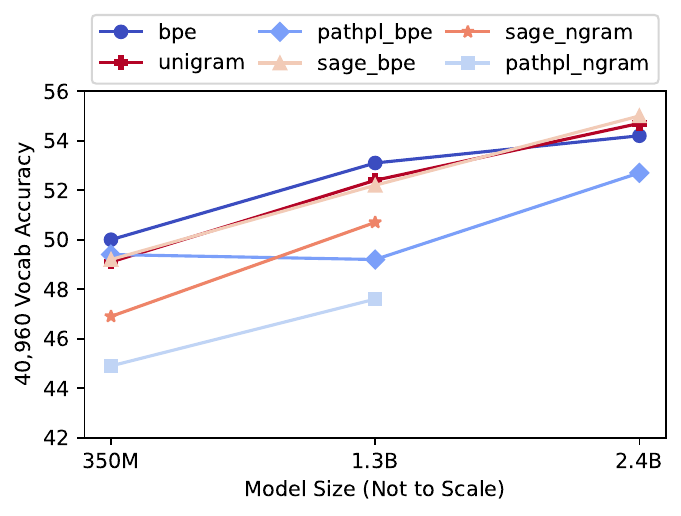}
    \caption{40,960 vocab average accuracy at various models sizes}
    \label{fig:larger_size}
\end{figure}

To examine the dependency on model size, we build larger models of 1.3B parameters for 6 of our experiments, and 2.4B parameters for 4 of them.  These models were trained over the same 200 billion tokens.  In the interest of computational time, these larger models were only run at a single vocabulary size of 40,960. The average results over the 10 task accuracies for these models is given in \autoref{fig:larger_size}. See \autoref{tab:summary1.2} in \autoref{app:experimental_results} for the numerical values.

It is noteworthy from the prevalence of crossing lines in \autoref{fig:larger_size} that the relative performance of the tokenizers do vary by model size, and that there is a group of tokenizers that are trading places being at the top for various model sizes. This aligns with our observation that the top 6 tokenizers were within the noise, and not significantly better than each other in the 350M models.

% \begin{figure}[ht]
%     \centering
%     \includegraphics[width=\columnwidth]{graphics/pieces_initvocab_pathpiece.pdf}
%     \caption{Initial vocab of PathPieceL.\\
%         Pairwise $p$-values between the pairs of runs are \\
%     $p$(1,9)=0.010, $p$(1,16)=3.5e-7, $p$(9,16)=8.5e-5.}
%     \label{fig:pieces_initvocab_pathpiece}
% \end{figure}

% \begin{figure}[ht]
%     \centering
%     \includegraphics[width=\columnwidth]{graphics/pieces_initvocab_sage.pdf}
%     \caption{Initial vocab of SaGe.\\
%         Pairwise $p$-values between the pairs of runs are \\
%     $p$(6,8)=0.019, $p$(6,10)=0.013, $p$(8,10)=0.15.}
%     \label{fig:pieces_initvocab_sage}
% \end{figure}

\section{Conclusion}
\label{sec:conc}
We investigate the hypothesis that reducing the corpus token count (CTC) would improve downstream performance, as suggested by \citet{galle-2019-investigating} and \citet{goldman2024unpacking} when they varied aspects of BPE.
When comparing CTC and downstream accuracy across all our experimental settings in \autoref{fig:total_tok_vs_accuracy}, we do not find a clear relationship between the two.  We expand on the findings of \citet{ali2024tokenizer} who did not find a strong relation when comparing 3 tokenizers, as we run 18 experiments varying the tokenizer, initial vocabulary, pre-tokenizer, and inference method. Our results suggest compression is not a straightforward explanation of what makes a tokenizer effective.

Finally, this work makes several practical contributions: (1) vocabulary size has little impact on downstream performance over the range of sizes we examined (\S\ref{subsec:vocabsize}); (2) five different tokenizers all perform comparably, with none outperforming at statistical significance (\S\ref{subsec:overall}); (3) BPE initial vocabularies work best for top-down vocabulary construction (\S\ref{subsec:initvocab}). To further encourage research in this direction, we make all of our trained vocabularies publicly available, along with the model weights from our 64 language models.

%We hope that these experiments will encourage further consideration of the decisions involved at each state of the tokenization process.

%\clearpage

\section*{Limitations}
\label{sec:limitations}

The objective of this work is to offer a comprehensive analysis of the tokenization process. However, our findings were constrained to particular tasks and models. Given the degrees of freedom, such as choice of downstream tasks, model, vocabulary size, etc., there is a potential risk of inadvertently considering our results as universally applicable to all NLP tasks; results may not generalize to other domains of tasks.

Additionally, our experiments were exclusively with English language text, and it is not clear how these results will extend to other languages. In particular, our finding that pre-tokenization is crucial for effective downstream accuracy is not applicable to languages without space-delimited words.

We conducted experiments for three district vocabulary sizes, and we reported averaged results across these experiments. With additional compute resources and time, it could be beneficial to conduct further experiments to gain a better estimate of any potential noise. For example, in \autoref{fig:checkpoint_graph} of \autoref{sec:model_convergence}, the 100k checkpoint at the 1.3B model size is worse than expected, indicating that noise could be an issue.

Finally, the selection of downstream tasks can have a strong impact on results. To allow for meaningful results, we attempted to select tasks that were neither too difficult nor too easy for the 350M parameter models, but other choices could lead to different outcomes. There does not seem to be a good, objective criteria for selecting a finite set of task to well-represent global performance.

\section*{Ethics Statement}

We have used the commonly used public dataset The Pile, which has not undergone a formal ethics review~\cite{DBLP:journals/corr/abs-2201-07311}. Our models may include biases from the training data. 

Our experimentation has used considerable energy. Each 350M parameter run took approximately 48 hours on (4) p4de nodes, each containing 8 NVIDIA A100 GPUs. We trained 62 models, including the 8 RandTrain runs in \autoref{app:randomtrain}. The (6) 1.3B parameters models took approximately 69 hours to train on (4) p4de nodes, while the (4) 2.4B models took approximately 117 hours to train on (8) p4de nodes. In total, training required 17,304 hours of p4de usage (138,432 GPU hours).

\section*{Acknowledgments}
Thanks to Charles Lovering at Kensho for his insightful suggestions, and to Michael Krumdick, Mike Arov, and Brian Chen at Kensho for their help with the language model development process.
This research was supported in part by the Israel Science Foundation (grant No. 1166/23).  Thanks to an anonymous reviewer who pointed out the large change in CTC when comparing Huggingface BPE and Unigram, in contrast to the previous literature using the SentencePiece implementations \cite{kudo-richardson-2018-sentencepiece}.

\bibliography{anthology,pathpiece}

\appendix
\section{Expanded description of \pathpiece{}}
\label{sec:full_pathpiece}

This section provides a self-contained explanation of \pathpiece{}, expanding on the one in \S\ref{sec:pathpiece}, with additional details on the vocabulary construction and complexity.  

In order to design an optimal vocabulary $\mathcal{V}$, it is first necessary to know how the vocabulary will be used to tokenize.  There can be no best vocabulary in the abstract.  Thus, we first present a new lossless subword tokenizer \pathpiece{}. This tokenization over our training corpus will provide the context to design a coherent vocabulary.

\subsection{Tokenization for a given vocabulary}

We work at the byte level, and require that all 256 single byte tokens are included in any given vocabulary $\mathcal{V}$. This avoids any out-of-vocabulary tokens by falling back to single bytes in the worst case.

Tokenization can be viewed as a compression problem, where we would like to tokenize text in a few tokens as possible. This has direct benefits, as it allows more text to fit in a given context window. A Minimum Description Length (MDL) argument can also be made that the tokenization using the fewest tokens best describes the data, although we saw in Subsection \ref{subsec:space_pre_tokenization} this may not always hold in practice.
    
Tokenizers such as BPE and WordPiece make greedy decisions, such as choosing which pair of current tokens to merge to create a new one, which results in tokenizations that may use more tokens than necessary. In contrast, \pathpiece{} will find an optimal tokenization by finding a shortest path through a Directed Acyclic Graph (DAG).
Informally, each byte $i$ of training data forms a node in the graph, and there is an edge if the $w$ byte sequence ending at $i$ is a token in $\mathcal{V}$.  

An implementation of \pathpiece{} is given in Algorithm \ref{alg:pathpiece2}, where input $d$ is a text document of $n$ bytes, $\mathcal{V}$ is a given vocabulary, and $L$ is a limit on the maximum width of a token in bytes. It has complexity $O(nL)$, following directly from the two nested \texttt{for}-loops.
It iterates over the bytes $i$ in $d$, computing 4 values for each.  It computes the shortest path length $pl[i]$ in tokens up to and including byte $i$, the width $wid[i]$ of a token with that shortest path length, and the solution count $sc[i]$ of optimal solutions found thus far with that shortest length.  We also remember the valid tokens of width 2 or more ending at each location $i$ in $vt[i]$, which will be used in the next section. 

There will be multiple tokenizations with the same optimal length, so some sort of tiebreaker is needed. The longest token or a randomly selected token are obvious choices.  We have presented the random tiebreaker method here, where a random solution is selected in a single pass in lines 29-32 of the listing using an idea from reservoir sampling \cite{Vitter1985}.

A backward pass through $d$ constructs the optimal tokenization from the $wid[e]$ values from the forward pass.

\begin{algorithm}[hbt!]
\caption{\pathpiece{} segmentation.}\label{alg:pathpiece2}
\begin{algorithmic}[1]
\small
\Procedure {PathPiece}{$d, \mathcal{V}, L$}
\State $n \leftarrow \len(d)$ 
\Comment{document length}
\For {$i \leftarrow 1, n$}
    \State $wid[i] \leftarrow 0$
    \Comment{shortest path token}
    \State $pl[i] \leftarrow \infty$
    \Comment{shortest path len}
    \State $sc[i] \leftarrow 0$
    \Comment{solution count}
    \State $vt[i] \leftarrow [\,]$
    \Comment{valid token list}
\EndFor
\For {$e \leftarrow 1, n$}
\Comment{token end}
    \For {$w \leftarrow 1, L$}
        \Comment{token width}
        \State $s \leftarrow e - w + 1$
        \Comment{token start}
    
        \If {$s \ge 1$}
        \Comment{$s$ in range}
            \State $t  \leftarrow d[s:e]$
            \Comment{token}
            \If {$t \in \mathcal{V}$}
                \If {$s = 1$}
                \Comment{1 tok path}
                    \State $wid[e] \leftarrow w$
                    \State $pl[e] \leftarrow 1$
                    \State $sc[e] \leftarrow 1$
                \Else
                    \If {$w \ge 2$} 
                        \State $vt[e].\append(w)$
                    \EndIf
                    \State $nl \leftarrow pl[s-1] + 1$ 
                    \If {$nl < pl[e]$}
                            \State $pl[e] \leftarrow nl$
                            \State $wid[e] \leftarrow w$
                            \State $sc[e] \leftarrow 1$                
                    \ElsIf {$nl = pl[e]$} 
                        \State{$sc[e] \leftarrow sc[e] + 1$}
                        \State $r \leftarrow \rand()$
                        \If {$r \le 1 / sc[e]$}
                            \State $wid[e] \leftarrow w$
                        \EndIf
                    \EndIf
                \EndIf
            \EndIf
        \EndIf    
    \EndFor
\EndFor
\State $T \leftarrow [\,]$
\Comment{output token list}
\State $e \leftarrow n$
\Comment{start at end of $d$}
\While {$e \ge 1$}
    \State $w \leftarrow wid[e]$
    \Comment{width of short path tok}
    \State $s \leftarrow e - w + 1$
    \Comment{token start}
    \State $t \leftarrow d[s:e]$
    \Comment{token}
    \State $T.\append(t)$
    \State $e \leftarrow  e - w$
    \Comment{back up a token}
\EndWhile
\State \textbf{return} {$\reversed(T)$}
\Comment{reverse order}
\EndProcedure
\end{algorithmic}
\end{algorithm}

\subsection{Optimal Vocabulary Construction}
\label{app:vocab}

\subsubsection{Vocabulary Initialization}

We will build an optimal vocabulary by starting from a large initial one, and sequentially omitting batches of tokens. We start with the most frequently occurring byte $n$-grams in a training corpus, of width 1 to $L$, or a large vocabulary trained by BPE or Unigram. We then add any single byte tokens that were not already included, making room by dropping the tokens with the lowest counts.
In our experiments we used an initial vocabulary size of $|\mathcal{V}| = 2^{18} = 262,144$.

\subsubsection{Increase from omitting a token}

Given a \pathpiece{} tokenization $t_1,\dots,t_{K_d}$, $\forall{d \in \mathcal{C}}$ for training corpus $\mathcal{C}$, we would like to know the increase in the overall length of a tokenization $K = \sum_d{K_d}$ from omitting a given token $t$ from our vocabulary, $\mathcal{V} \setminus \{t\}$ and recomputing the tokenization. Tokens with a low increase are good candidates to remove from the vocabulary~\cite{kudo-2018-subword}.
However, doing this from scratch for each $t$ would be a very expensive $O(nL|\mathcal{V}|)$ operation. 

We make a simplifying assumption that allows us to compute these increases more efficiently.  We omit a specific token $t_k$ in the tokenization of document $d$, and compute the minimum increase $MI_{kd}$ in $K_d$ from not having that token $t_k$ in the tokenization of $d$.  We then aggregate over the documents to get the overall increase for $t$:
\begin{equation}
    MI_t = \sum_{d \in \mathcal{C}} \sum_{k=1 | t_k = t}^{K_d} MI_{kd}.
    \label{eq:aggregation}
\end{equation}
This is similar to computing the increase from $\mathcal{V} \setminus \{t\}$, but ignores interaction effects from having several occurrences of the same token $t$ close to each other in a given document. 

With \pathpiece{}, it turns out we can compute the minimum increase in tokenization length without actually recomputing the tokenization. Any tokenization not containing $t_k$ must either contain a token boundary somewhere inside of $t_k$ breaking it in two, or it must contain a token that entirely contains $t_k$ as a superset. Our approach will be to enumerate all the occurrences for these two cases, and to find the minimum increase $MI_{kd}$ overall.

Before considering these two cases, there is a shortcut that often tells us that there would be no increase due to omitting $t_k$ ending at index $e$.  We computed the solution count vector $sc[e]$ when running Algorithm \ref{alg:pathpiece2}.  If $sc[e] > 1$ for a token ending at $e$, then the backward pass could simply select one of the alternate optimal tokens, and find an overall tokenization of the same length.  

Let $t_k$ start at index $s$ and end at index $e$, inclusive. Remember that path length $pl[i]$ represents the number of tokens required for shortest path up to and including byte $i$.  We can also run Algorithm \ref{alg:pathpiece2} backwards on $d$, computing a similar vector of backwards path lengths $bpl[i]$, representing the number of tokens on a path from the end of the data up to and including byte $i$.  The overall minimum length of a tokenization with a token boundary after byte $j$ is thus:
\begin{equation}
    K_j^b = pl[j] + bpl[j+1].
\end{equation}
We have added an extra constraint on the shortest path, that there is a break at $j$, so clearly $K_j^{br} \ge pl[n]$. The minimum increase for the case of having a token boundary within $t_k$ is thus:
\begin{equation}
    MI_{kd}^b = \min_{j=s,\dots,e-1} {K_j^b - pl[n]}.
\end{equation}
Each token $t_k$ will have no more than $L-1$ potential internal breaks, so the complexity of computing $MI_{kd}^{b}$ is $O(L)$.

The minimum increase from omitting $t_k$ could also be on a tokenization containing a strict superset of $t_k$.  Let this superset token be $t_k'$, with start $s'$ and end $e'$ inclusive.  To be a strict superset jumping over $t_k$, we must have $s' < s$ and $e' \ge e$, or $s' \le s$ and $e' > e$, subject to the constraint that the width $w' = e' - s' + 1 \le L$.  In this case, the minimum length of using the superset token $t_k'$ would be:
\begin{equation}
    K_{t_k'}^s = pl[s'-1] + bpl[e'+1] + 1,
\end{equation}
which is the path length to get to the byte before $t_k'$, plus the path length go backwards to the byte after $t_k'$, plus 1 for the token $t_k'$ itself.

We remembered a list of the widths of the tokens ending at each byte, $vt[e]$ in Algorithm \ref{alg:pathpiece2}.  The set of superset tokens $S$ can be found by examining the $O(L)$ potential $e'$, and then seeing if the $w' \in vt[e']$ give tokens forming a strict superset.  There are $O(L)$ potential tokens ending at $e'$ in $vt[e']$, so the overall complexity of finding the superset tokens is $O(L^2)$

Similar to the previous case, we can compute the minimum increase from replacing $t_k$ with a superset token by taking the minimum increase over the superset tokens:
\begin{equation}
    MI_{kd}^s = \min_{t_k' \in S} {K_{t_k'}^s - pl[n]}.
\end{equation}

Finally, the overall minimum increase $MI_{kd}$ from omitting $t_k$ is simply
\begin{equation}
    MI_{kd} = \min(MI_{kd}^b,MI_{kd}^s).
\end{equation}

When aggregating over all $t_k$ according to eq (\ref{eq:aggregation}), one iteration of the vocabulary construction procedure will have complexity $O(n L^2)$.

\section{Language Model Parameters}
\label{section:lang_model_params}

The 350M parameter models were trained using the MPT architecture\footnote{https://github.com/mosaicml/llm-foundry} with the following parameters:

\begin{Verbatim}[fontsize=\small]
# Model
model:
  name: mpt_causal_lm
  init_deice: meta
  d_model: 1024
  n_heads: 16
  n_layers: 24
  expansion_ratio: 4
  max_seq_len: 2048
  attn_config:
    alibi: true
    attn_impl: triton
    clip_qkv: 6

# Optimization
device_eval_batch_size: 5
device_train_microbatch_size: 32
global_train_batch_size: 1024 # ~2M tokens
max_duration: 100000ba # ~200B tokens

optimizer:
  name: decoupled_adamw
  lr: 3.0e-4
  betas:
  - 0.9
  - 0.95
  eps: 1.0e-08
  weight_decay: 0.0001

scheduler:
  name: cosine_with_warmup
  t_warmup: 0.05dur
  alpha_f: 0.1

# System
precision: amp_bf16

# Algos and Callbacks
algorithms:
  gradient_clipping:
    clipping_threshold: 1
    clipping_type: norm

\end{Verbatim}

The 1.3B parameter models simply changes:

\begin{Verbatim}[fontsize=\small]
d_model: 1024
\end{Verbatim}

The 2.4B parameter models updates:

\begin{Verbatim}[fontsize=\small]
d_model: 2560
n_heads: 20
n_layers: 32

\end{Verbatim}

\section{Description of Downstream Tasks}
\label{section:descr_benchmark_tasks}

To evaluate the performance of our various tokenization experiments, we select ten competitive benchmarks from \texttt{lm-evaluation-harness} \cite{eval-harness}\footnote{https://github.com/EleutherAI/lm-evaluation-harness}, that we broadly categorize into three types of Question Answering (QA) tasks: Knowledge-based, Common-sense Reasoning and Context-based.

\textbf{Knowledge Based Tasks}
Knowledge based tasks in this study expect LLMs to answer questions based on domain-specific internal retrieval. Our Knowledge-based baselines in this work include:

\textit{SciQ}: The SciQ task, proposed by \citet{sciq} contains a total of 13,679 science exam questions. The questions are in multiple-choice format with 4 answer options each. An additional text is provided as supporting evidence for a majority of the answers.

\textit{ARC (AI2 Reasoning Challenge)}: \citet{arc_easy} compiles grade-school level, multiple-choice science question dataset consists of 7,787 science exam questions that are split into “easy” and “hard” sets. For this study, we employ the easy set of 5,197 questions, each having 4 answer choices.

\textit{MathQA}: \citet{mathqa} introduce a dataset of math word problems that require LLMs to use their internal understanding of mathematical equations and arithmetic comprehension. Similar to SciQ, this dataset consists of 37k multiple-choice questions with the equations for each used annotated.

\textit{HendrycksTest}: \citet{hendrycksTests} provide a comprehensive suite of of multiple-choice tests for assessing text models in multi-task contexts. It comprises of 57 tasks such as elementary mathematics, US history, law of which we use the sociology and marketing tests.

\textbf{Commonsense Reasoning Tasks}
These tasks assess the model's capability to infer and reason about everyday scenarios based on implicit knowledge.

\textit{COPA (Choice of Plausible Alternatives)}: COPA proposed by \citet{copa} is a benchmark for assessing progress in open-domain commonsense causal reasoning. It consists of 1000 questions where each question is composed of a premise and two alternatives. The task is to select the alternative that more plausibly has a causal relation with the premise.

\textit{PiQA (Physical Interaction Question Answering)}: \citet{piqa} introduce a task that assess the understanding of physical commonsense by language models. Comprised of everyday situation with a preference for atypical solutions, this dataset is formulated as multiple choice question with two possible solutions choices for each question.

\textit{Winograd Schema Challenge}: \citet{winograd} define a task with a pair of sentences that differ only in one or two words and that contain a referential ambiguity that is resolved in opposite directions in the two sentences. This dataset of 273 tasks test language model understanding of the content of the text and disambiguation ability.

\textbf{Context Based Tasks}
These tasks are reliant on understanding context and drawing conclusions from it.

\textit{RACE (Reading Comprehension from Examinations)}: RACE proposed by \citet{race} is a collection of English questions set aside to Chinese school students. Each item is divided into two parts, a passage that the student must read and a set of 4 potential answers, requiring extraction and reasoning capabilities.\par

\textit{QA4MRE (Question Answering for Machine Reading Evaluation)}: QA4MRE by \citet{qa4mre} is a benchmark designed to resolve reading comprehension challenges. This task focuses on reading of single documents and identifying the answers to a set of questions. Questions are in the form of multiple choice with one correct option.

Our goal was to select tasks where a 350M parameter model could do significantly better than random chance, avoiding evaluation right at the noisier random threshold. We started with the tasks that had a non-zero random score (indicating multiple choice), and then chose tasks where BPE at a vocabulary size 40,960 could do well above random.  In the end, the average accuracy across models was more than 15\% above random on all tasks. 

Note that in results tables we have shortened the name hendrycksTest-marketing to marketing, hendrycksTest-sociology to sociology, and qa4mre\_2013 to qa4mre.

\section{Effect of model convergence}
\label{sec:model_convergence}

\begin{figure}[ht]
    \centering
    \includegraphics[width=7.5cm]{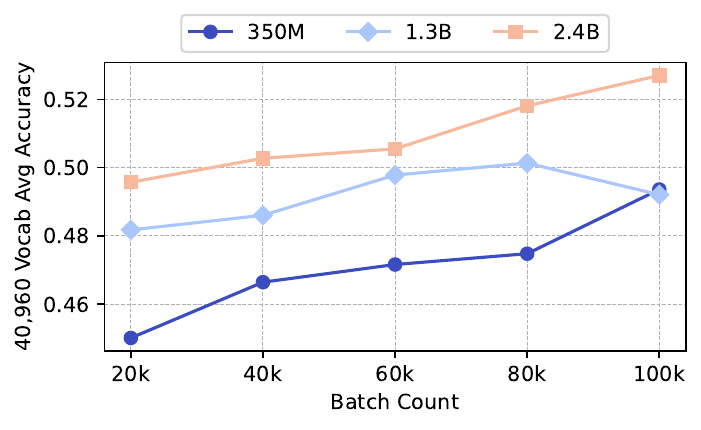}
    \caption{Checkpoint accuracy values for PathPieceL with an initial vocabulary from BPE and a vocabulary size of 40,960, evaluated at 5 checkpoints.}
    \label{fig:checkpoint_graph}
\end{figure}

Each model was trained on around 200 billion tokens. \autoref{fig:checkpoint_graph} gives a plot of the average accuracy for PathPieceL with a BPE initial vocabulary and a vocabulary size of 40,960 at various checkpoints in the language model training process. It also shows checkpoints for the larger 1.3B and 2.4B models discussed in the Limitations section. With the exception of the 100k checkpoint at 1.3B, the model appears to be continually improving. It is unclear why the 100k checkpoint did so poorly.

\section{Additional Analysis}
\label{app:additional_analysis}

Here we additional details for results from \S\ref{sec:analysis} that are just summarized in the text in the interest of space. 

\subsection{Segmentation}
\label{sub:segmentation}

Tokenizers often use the segmentation strategy that is used in vocabulary construction.
However, any vocabulary can also be used with \pathpiece{} and with the greedy left-to-right segmentation methods.

We find that BPE works quite well with greedy segmentation (overall rank 4, insignificantly different from the top rank), but not with the shortest-path segmentation of \textsc{PathPieceL} (13).

\begin{figure}[ht]
    \centering
    \includegraphics[width=\columnwidth]{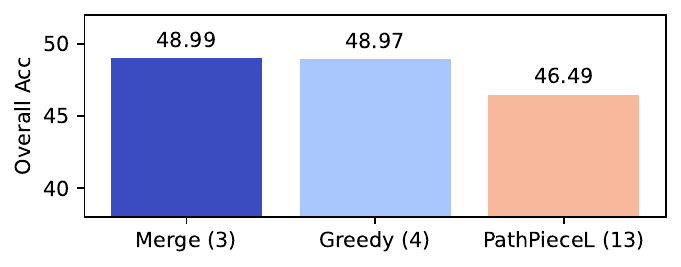}
    \captionsetup{justification=centering}
    \caption{Segmentation of BPE.\\
        Pairwise $p$-values between the pairs of runs are \\
    $p$(3,4)\-=\-0.52, $p$(3,13)=4.4e-5, $p$(4,13)=8.8e-6.}
    \label{fig:pieces_seg_bpe}
\end{figure}

Unigram, on the other hand, seems to be more tightly tied to its default maximum likelihood segmentation (2), which was significantly better than both Greedy (7) and \textsc{PathPieceL} (17).

\begin{figure}[!ht]
    \centering
    \includegraphics[width=\columnwidth]{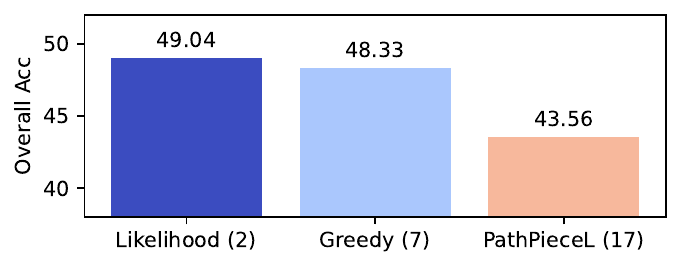}
    \captionsetup{justification=centering}
    \caption{Segmentation of Unigram.\\
        Pairwise $p$-values between the pairs of runs are \\
    $p$(2,7)=0.041, $p$(2,17)=2.9e-06, $p$(7,17)=2.9e-06}  % 
    \label{fig:pieces_seg_unigram}
\end{figure}

\subsection{Digit Pre-tokenization}

\begin{figure}[ht]
    \centering
    \includegraphics[width=\columnwidth]{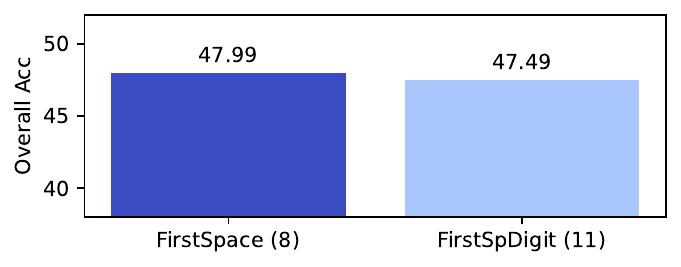}
    \captionsetup{justification=centering}
    \caption{Pre-tokenization of Sage, $n$-gram initial,\\$p$=0.025.}
    \label{fig:pieces_digit_pretok_sage}
\end{figure}

\begin{figure}[ht]
    \centering
    \includegraphics[width=\columnwidth]{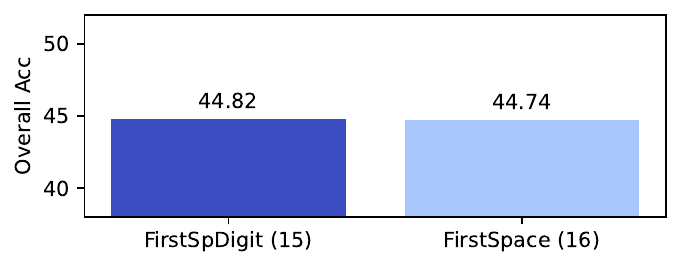}
    \captionsetup{justification=centering}
    \caption{Pre-tokenization of PathPieceL $n$-gram,\\$p$=0.54.}
    \label{fig:pieces_digit_pretok_2}
\end{figure}

We have two examples isolating Digit pre-token\-iza\-tion, when a digit must always be its own token.   \autoref{fig:pieces_digit_pretok_sage} shows Digit hurts for Sage with an $n$-gram initial vocabulary, while \autoref{fig:pieces_digit_pretok_2} shows no significant differences for PathPieceL, also with an $n$-gram initial vocabulary.

With the exception of mathqa, none of our downstream tasks were particularly mathematical in nature. It is likely this makes it hard to make a definitive judgement on Digit with our experiments.  

\subsection{Vocabulary Construction}
\label{sub:venn_unigram}

\begin{figure}[ht]
    \centering
    \includegraphics[width=\columnwidth]{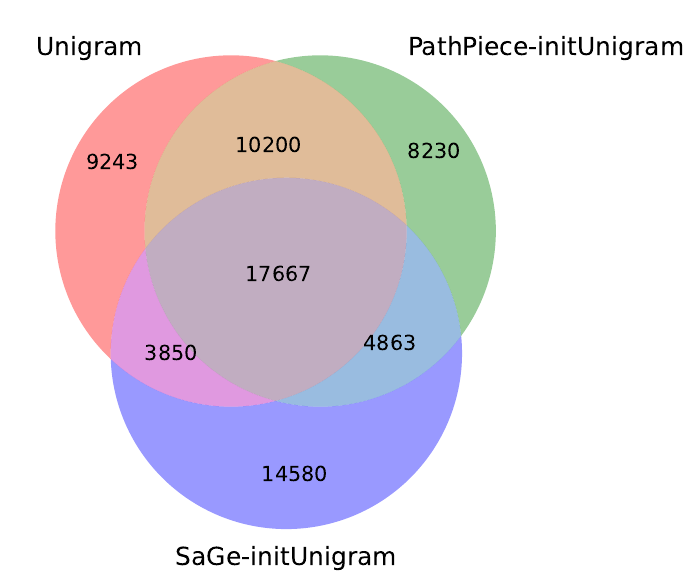}
    \captionsetup{justification=centering}
    \caption{Venn diagrams comparing 40,960 token vocabularies of Unigram, PathPieceL and SaGe, where the latter two were both trained from a initial Unigram vocabulary of size 262,144}
    \label{fig:venn_diagram_unigram}
\end{figure}

\autoref{fig:venn_diagram_unigram} gives a Venn diagram of the overlap in vocabularies between Unigram, PathPieceL, and SaGe, when both PathPieceL and SaGe were constructed from a large initial vocabulary of size 262,144 from Unigram.  As with \autoref{fig:venn_diagram_bpe}, we see that PathPiece is more similar to Unigram, while SaGe chose more distinct tokens.

\subsection{PathPiece tie breaking}

\begin{figure}[ht]
    \centering
    \includegraphics[width=\columnwidth]{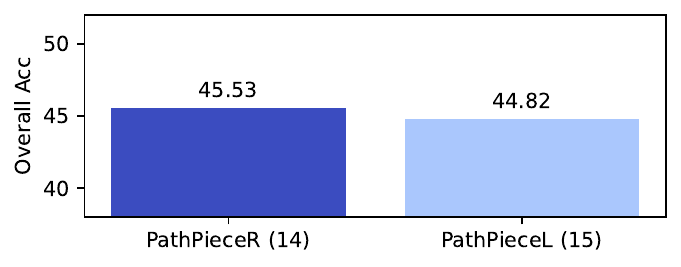}
    \captionsetup{justification=centering}
    \caption{Tiebreaking PathPieceL vs PathPieceR\\with $n$-gram, $p$=0.067.}
    \label{fig:pieces_tiebreaker}
\end{figure}

The difference in tie breaking between choosing the longest token with PathPieceL versus choosing randomly with PathPieceR turns out not to be significant, as seen in in \autoref{fig:pieces_tiebreaker}.

\section{RandTrain}
\label{app:randomtrain}

None of our experiments completely isolate the effect of the vocabulary construction step. We created a new baseline random vocabulary construction approach, RandTrain, in an attempt to do so.  It is meant to work with a top-down method like SaGe or PathPieceL, and uses the same initial vocabulary, pre-tokenization, and segmentation as either of those, with a simple vocabulary construction algorithm.  

We compute a count for each token in the vocabulary.  For the top $n$-gram initial vocabulary it is simply the $n$-gram count from the training corpus.  For a BPE initial vocabulary we tokenized the training corpus with BPE and the large initial vocabulary, and then use the occurrence counts of each token.  We  normalize these counts into target selection probabilities $p_k$ for token $t_k$.

The RandTrain vocabulary construction process is simply to randomly sample our desired vocabulary size $m$ of tokens from the initial vocabulary, proportionally to $p_k$, without replacement. Sampling without replacement is necessary to avoid have duplicate words in the vocabulary. Interestingly, this is not possible if there are any $p_k > 1/m$, which are termed infeasible or overweight items~\cite{DBLP:journals/corr/abs-1012-0256}. The intuition behind this is when selecting $m$ items without replacement, it is not possible to select a given item more than once.  So even if an item is always selected in a sample, the selection probability will be $p_k = 1/m$.

We sampled without replacement using the A-ES Algorithm described in \citet{DBLP:journals/corr/abs-1012-0256}.  A significant number the most common tokens in the vocabulary were infeasible and hence were unable to reach their target $p_k$.  A token with a higher $p_k$ is more likely to be sampled than a token with a lower one, but they may significantly differ from their target $p_k$.

We build 6 RandTrain models with 3 different types of pre-tokenization, and with Greedy segmentation to compare to SaGe, and PathPieceL segmentation to compare to PathPieceL. We only used a single vocabulary size of 40,960, so $p$-values are only computed on the 10 task accuracies, rather than the 30 used elsewhere. Task level accuracies are given in \autoref{tab:350_40960_pt1} and \autoref{tab:350_40960_pt2} in \autoref{app:experimental_results}.

\begin{figure}[ht]
    \centering
    \includegraphics[width=\columnwidth]{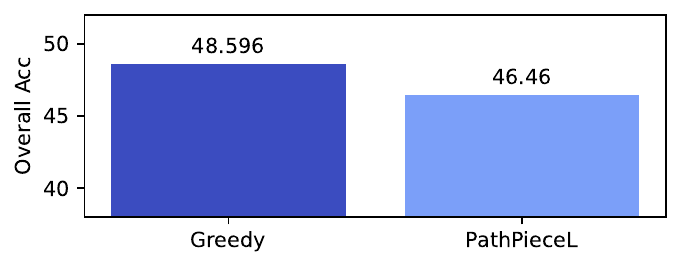}
    \captionsetup{justification=centering}
    \caption{Comparison of Greedy and PathPieceL segmentation, with RandTrain vocabulary construction, BPE initial vocab, and FirstSpace pre-tokenization, $p$=0.0273}
    \label{fig:randtrain_Rand_BPE_FSp}
\end{figure}

Before comparing RandTrain to SaGe and PathPieceL, we will compare our RandTrain runs to each other, with different segmentation approaches. In \autoref{fig:randtrain_Rand_BPE_FSp} and \autoref{fig:randtrain_Rand_ngram_FSpD} we have pairs of RandTrain runs that only vary by the segmentation method. 

% randtrain_Rand_BPE_FSp.pdf
% randtrain_Rand_ngram_FSp.pdf
% randtrain_Rand_ngram_FSpD.pdf

\begin{figure}[ht]
    \centering
    \includegraphics[width=\columnwidth]{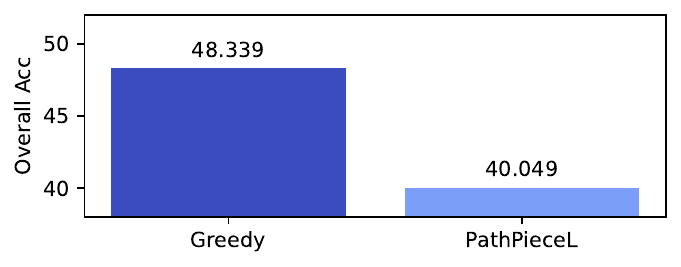}
    \captionsetup{justification=centering}
    \caption{Comparison of Greedy and PathPieceL segmentation, with RandTrain vocabulary construction, $n$-gram initial vocab, and FirstSpace pre-tokenization, $p$=0.00195}
    \label{fig:randtrain_Rand_ngram_FSp}
\end{figure}

\begin{figure}[ht]
    \centering
    \includegraphics[width=\columnwidth]{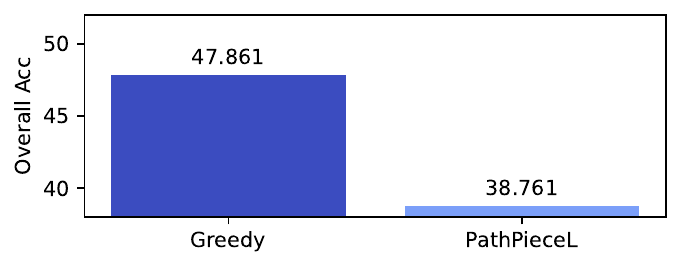}
    \captionsetup{justification=centering}
    \caption{Comparison of Greedy and PathPieceL segmentation, with RandTrain vocabulary construction, $n$-gram initial vocab, and FirstSpaceDigit pre-tokenization, $p$=0.00293}
    \label{fig:randtrain_Rand_ngram_FSpD}
\end{figure}

In line with Subsection \ref{sub:segmentation}, Greedy performs significantly better than PathPieceL segmentation in all 3 cases. However, for the two cases with an $n$-gram initial vocabulary the PathPieceL segmentation did extremely poorly. The RandTrain vocabulary construction, $n$-gram initial vocabulary, and PathPieceL segmentation interact somehow to give accuracies well below any others.  

\begin{figure}[ht]
    \centering
    \includegraphics[width=\columnwidth]{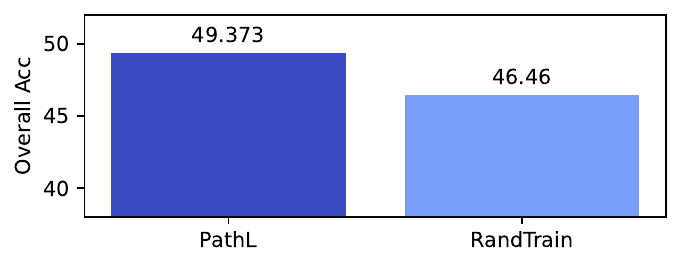}
    \captionsetup{justification=centering}
    \caption{Comparison of PathPieceL and RandTrain, with BPE initial vocab, and FirstSpace pre-tokenization, $p$=0.0137}
    \label{fig:randtrain_BPE_FSp_Greedy_PathLvsRand}
\end{figure}

This makes the comparison of RandTrain to PathPieceL less informative.  We can see in \autoref{fig:randtrain_BPE_FSp_Greedy_PathLvsRand} that PathPieceL is significantly better than RandTrain with a BPE initial vocabulary. 

\begin{figure}[ht]
    \centering
    \includegraphics[width=\columnwidth]{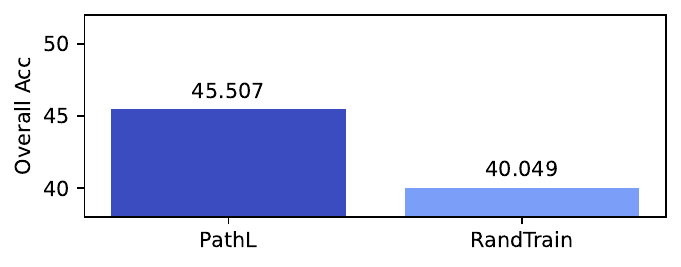}
    \captionsetup{justification=centering}
    \caption{Comparison of PathPieceL and RandTrain, with $n$-gram initial vocab, and FirstSpace pre-tokenization, $p$=9.77e-4}
    \label{fig:randtrain_ngram_FSp_Greedy_PathLvsRand}
\end{figure}

\begin{figure}[ht]
    \centering
    \includegraphics[width=\columnwidth]{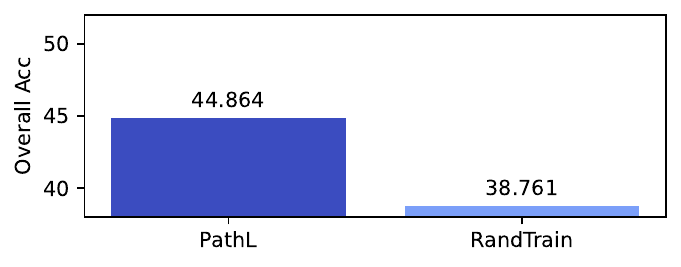}
    \captionsetup{justification=centering}
    \caption{Comparison of PathPieceL and RandTrain, with $n$-gram initial vocab, and FirstSpaceDigits pre-tokenization, $p$=0.00977}
    \label{fig:randtrain_ngram_FSpD_Greedy_PathLvsRand}
\end{figure}

% randtrain_BPE_FSp_Greedy_PathLvsRand.pdf
% randtrain_ngram_FSp_Greedy_PathLvsRand.pdf
% randtrain_ngram_FSpD_Greedy_PathLvsRand.pdf
However, the other two comparisons in \autoref{fig:randtrain_ngram_FSp_Greedy_PathLvsRand} are \autoref{fig:randtrain_ngram_FSpD_Greedy_PathLvsRand} are not that meaningful.  They are significantly better, but that is more about the weak baseline of RandTrain with PathPieceL segmentation than anything positive about PathPieceL.  

\begin{figure}[ht]
    \centering
    \includegraphics[width=\columnwidth]{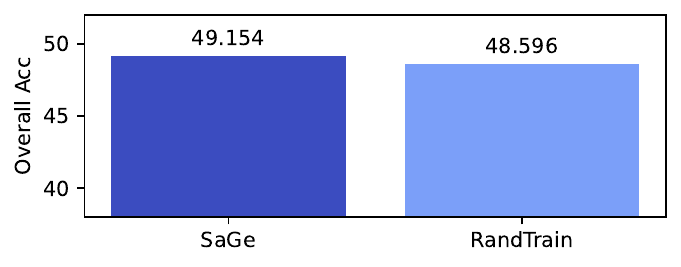}
    \captionsetup{justification=centering}
    \caption{Comparison of SaGe and RandTrain, with BPE initial vocab, and FirstSpace pre-tokenization, $p$=0.0645}
    \label{fig:randtrain_BPE_FSp_Greedy}
\end{figure}

\begin{figure}[ht]
    \centering
    \includegraphics[width=\columnwidth]{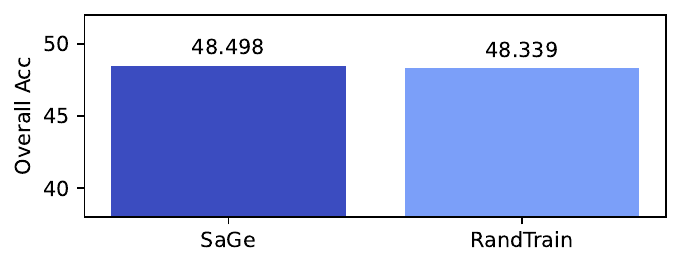}
    \captionsetup{justification=centering}
    \caption{Comparison of SaGe and RandTrain, with $n$-gram initial vocab, and FirstSpace pre-tokenization, $p$=0.688}
    \label{fig:randtrain_ngram_FSp_Greedy}
\end{figure}

The remaining comparison between SaGe and RandTrain is more interesting. In \autoref{fig:randtrain_BPE_FSp_Greedy} and \autoref{fig:randtrain_ngram_FSp_Greedy} SaGe was not significantly better than RandTrain, with a $p$-value of 0.0645.

% randtrain_BPE_FSp_Greedy.pdf
% randtrain_ngram_FSp_Greedy.pdf
% randtrain_ngram_FSpD_Greedy.pdf

\begin{figure}[ht]
    \centering
    \includegraphics[width=\columnwidth]{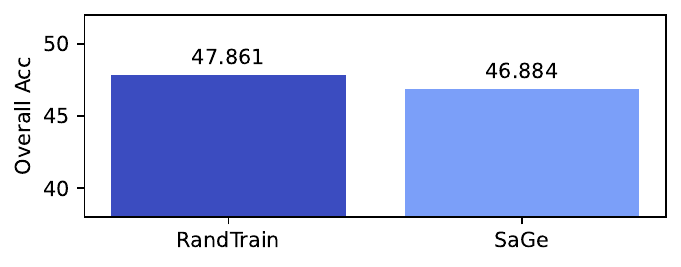}
    \captionsetup{justification=centering}
    \caption{Comparison of RandTrain and SaGe, with $n$-gram initial vocab, and FirstSpaceDigit pre-tokenization, $p$=0.15}
    \label{fig:randtrain_ngram_FSpD_Greedy}
\end{figure}

The cases is even worse for the two $n$-gram initial vocabulary cases.  In \autoref{fig:randtrain_ngram_FSp_Greedy} the $p$-value was a 0.688, and in \autoref{fig:randtrain_ngram_FSpD_Greedy} RandTrain was actually better, although not significantly. 

We saw in \autoref{tab:overallavg_grouped} that both PathPieceL-BPE and SaGe-BPE are effective tokenizers.  In attempting to isolate the benefit from the vocabulary construction step, we see that PathPieceL-BPE outperforms our simple baseline.  However, SaGe was unable to outperform the baseline, perhaps implying that RandTrain may actually be a simple but fairly effective vocabulary construction method.

\section{Detailed Experimental Results}
\label{app:experimental_results}

This section gives the detailed accuracy results for the 10 downstream evaluation tasks on each model that was trained.  The tables are divided by the vocabulary size used, with \autoref{tab:350_32768_pt1} and \autoref{tab:350_32768_pt2} for 32,768; \autoref{tab:350_40960_pt1}
 and \autoref{tab:350_40960_pt2} for 40,960; and \autoref{tab:350_49152_pt1} and \autoref{tab:350_49152_pt2} for 49,152.  The highest value or values (in the case of ties) are shown in bold. \autoref{tab:overallavg} show the same results as \autoref{tab:overallavg_grouped}, but are sorted from best to worst by rank.
The corpus token count (CTC), R\'enyi efficiencies, and average accuracies for the 54 runs in \autoref{fig:total_tok_vs_accuracy} are given in \autoref{tab:fig3data}.

The detailed accuracy results for our 1.3B parameter models, which were all performed at a single vocabulary size of 40,960, are given in \autoref{tab:1.2_pt1} and \autoref{tab:1.2_pt2}. Average accuracy results for larger models of 1.3B and 2.4B parameters are given in \autoref{tab:summary1.2}.  See \S\ref{sec:limitations} for more discussion of this table.

\begin{table*}[hbt!]
    \centering
    \small
    \begin{tabular}{llllrrrrrr}
    \toprule
\textbf{Vocab Constr}	&	\textbf{Init Voc}	&	\textbf{Pre-tok}	&	\textbf{Segment}	&	\textbf{Avg}	&	\textbf{arc\_easy}	&	\textbf{copa}	&	\textbf{mktg}	&	\textbf{mathqa}	&	\textbf{piqa}	\\
\midrule

\multirow{3}{*}{BPE}	&		&	FirstSpace	&	Merge	&	48.8	&	51.2	&	69.0	&	32.9	&	\textbf{23.9}	&	66.3	\\
	&		&	FirstSpace	&	Greedy	&	48.3	&	51.9	&	66.0	&	32.9	&	23.7	&	65.6	\\
	&		&	FirstSpace	&	PathPieceL	&	45.6	&	45.6	&	61.0	&	29.9	&	23.0	&	60.5	\\
\midrule

\multirow{3}{*}{Unigram}	&		&	FirstSpace	&	Likelihood	&	49.2	&	50.7	&	73.0	&	30.8	&	23.1	&	66.3	\\
	&		&	FirstSpace	&	Greedy	&	47.9	&	50.3	&	68.0	&	31.2	&	23.1	&	65.2	\\
	&		&	FirstSpace	&	PathPieceL	&	43.6	&	41.2	&	57.0	&	31.6	&	22.0	&	60.6	\\
\midrule

WordPiece	&		&	FirstSpace	&	Greedy	&	48.5	&	\textbf{52.5}	&	64.0	&	32.5	&	\textbf{23.9}	&	65.6	\\
\midrule

\multirow{4}{*}{SaGe}	&	BPE	&	FirstSpace	&	Greedy	&	47.9	&	49.7	&	67.0	&	26.5	&	23.2	&	65.9	\\
	&	$n$-gram	&	FirstSpDigit	&	Greedy	&	48.4	&	50.3	&	71.0	&	29.5	&	22.0	&	65.1	\\
	&	$n$-gram	&	FirstSpace	&	Greedy	&	47.5	&	48.8	&	64.0	&	29.5	&	23.0	&	\textbf{66.6}	\\
	&	Unigram	&	FirstSpace	&	Greedy	&	48.4	&	52.0	&	\textbf{74.0}	&	27.8	&	22.7	&	65.7	\\
\midrule

\multirow{4}{*}{PathPieceL}	&	BPE	&	FirstSpace	&	PathPieceL	&	\textbf{49.3}	&	50.8	&	68.0	&	\textbf{34.2}	&	23.0	&	66.4	\\
	&	$n$-gram	&	FirstSpace	&	PathPieceL	&	44.8	&	42.3	&	61.0	&	27.4	&	23.0	&	61.2	\\
	&	$n$-gram	&	FirstSpDigit	&	PathPieceL	&	44.6	&	42.3	&	62.0	&	31.2	&	22.8	&	61.2	\\
	&	Unigram	&	FirstSpace	&	PathPieceL	&	46.9	&	50.4	&	64.0	&	24.8	&	23.5	&	66.2	\\
\midrule 

\multirow{3}{*}{PathPieceR}	&	$n$-gram	&	FirstSpDigit	&	PathPieceR	&	45.3	&	46.9	&	67.0	&	26.9	&	22.4	&	59.9	\\
	&	$n$-gram	&	None	&	PathPieceR	&	43.5	&	42.5	&	65.0	&	26.1	&	22.8	&	61.7	\\
	&	$n$-gram	&	SpaceDigit	&	PathPieceR	&	47.5	&	48.6	&	68.0	&	32.9	&	23.3	&	65.0	\\
\midrule

Random	&		&		&		&	32.0	&	25.0	&	50.0	&	25.0	&	20.0	&	50.00	\\

    \bottomrule
    \end{tabular}
   \caption{350M parameter model, 32,768 token vocabulary, accuracy (\%) on average and initial 5 tasks}
    \label{tab:350_32768_pt1}
    
\end{table*}

\begin{table*}[hbt!]
    \centering
    \small
    \begin{tabular}{llllrrrrr}
    \toprule
\textbf{Vocab Constr}	&	\textbf{Init Voc}	&	\textbf{Pre-tok}	&	\textbf{Segment}	&	\textbf{qa4mre}	&	\textbf{race}	&	\textbf{sciq}	&	\textbf{sociology}	&	\textbf{wsc273}	\\
\midrule

\multirow{3}{*}{BPE}	&		&	FirstSpace	&	Merge	&	29.6	&	29.2	&	87.3	&	30.9	&	67.8	\\
	&		&	FirstSpace	&	Greedy	&	27.5	&	30.7	&	88.0	&	30.9	&	66.3	\\
	&		&	FirstSpace	&	PathPieceL	&	28.2	&	29.0	&	83.8	&	28.4	&	66.3	\\
\midrule

\multirow{3}{*}{Unigram}	&		&	FirstSpace	&	Likelihood	&	31.0	&	30.2	&	86.4	&	31.8	&	\textbf{68.5}	\\
	&		&	FirstSpace	&	Greedy	&	28.9	&	30.6	&	86.9	&	31.8	&	62.6	\\
	&		&	FirstSpace	&	PathPieceL	&	29.9	&	27.5	&	74.6	&	26.4	&	65.6	\\
\midrule

WordPiece	&		&	FirstSpace	&	Greedy	&	\textbf{32.0}	&	30.7	&	88.5	&	27.9	&	67.4	\\
\midrule

\multirow{4}{*}{SaGe}	&	BPE	&	FirstSpace	&	Greedy	&	31.7	&	30.2	&	\textbf{89.0}	&	28.4	&	67.8	\\
	&	$n$-gram	&	FirstSpDigit	&	Greedy	&	31.0	&	30.3	&	86.6	&	32.3	&	66.0	\\
	&	$n$-gram	&	FirstSpace	&	Greedy	&	30.0	&	31.0	&	87.8	&	25.9	&	\textbf{68.5}	\\
	&	Unigram	&	FirstSpace	&	Greedy	&	29.6	&	28.9	&	88.2	&	32.3	&	63.0	\\
\midrule

\multirow{4}{*}{PathPieceL}	&	BPE	&	FirstSpace	&	PathPieceL	&	28.5	&	\textbf{31.1}	&	88.8	&	\textbf{35.3}	&	67.0	\\
	&	$n$-gram	&	FirstSpace	&	PathPieceL	&	30.3	&	27.3	&	80.0	&	32.8	&	62.6	\\
	&	$n$-gram	&	FirstSpDigit	&	PathPieceL	&	27.8	&	25.5	&	79.2	&	31.3	&	62.6	\\
	&	Unigram	&	FirstSpace	&	PathPieceL	&	29.6	&	30.6	&	87.6	&	24.4	&	68.1	\\
\midrule

\multirow{3}{*}{PathPieceR}	&	$n$-gram	&	FirstSpDigit	&	PathPieceR	&	28.5	&	29.4	&	78.6	&	28.9	&	64.5	\\
	&	$n$-gram	&	None	&	PathPieceR	&	27.1	&	27.0	&	77.7	&	28.9	&	56.0	\\
	&	$n$-gram	&	SpaceDigit	&	PathPieceR	&	25.0	&	29.4	&	85.7	&	32.3	&	64.8	\\
\midrule

Random	&		&		&		&	25.0	&	25.0	&	25.0	&	25.0	&	50.0	\\

    \bottomrule
    \end{tabular}
   \caption{350M parameter model, 32,768 token vocabulary, accuracy (\%) on remaining 5 tasks}
    \label{tab:350_32768_pt2}
    
\end{table*}

\begin{table*}[hbt!]
    \centering
    \small
    \begin{tabular}{llllrrrrrr}
    \toprule
\textbf{Vocab Constr}	&	\textbf{Init Voc}	&	\textbf{Pre-tok}	&	\textbf{Segment}	&	\textbf{Avg}	&	\textbf{arc\_easy}	&	\textbf{copa}	&	\textbf{mktg}	&	\textbf{mathqa}	&	\textbf{piqa}	\\
\midrule

\multirow{3}{*}{BPE}	&		&	FirstSpace	&	Merge	&	\textbf{50.0}	&	\textbf{52.7}	&	70.0	&	31.6	&	24.3	&	66.9	\\
	&		&	FirstSpace	&	Greedy	&	49.1	&	52.3	&	66.0	&	27.4	&	22.9	&	\textbf{66.9}	\\
	&		&	FirstSpace	&	PathPieceL	&	46.7	&	48.0	&	58.0	&	27.4	&	23.4	&	62.1	\\
\midrule

\multirow{3}{*}{Unigram}	&		&	FirstSpace	&	Likelihood	&	49.1	&	51.4	&	\textbf{71.0}	&	32.1	&	23.4	&	66.1	\\
Unigram	&		&	FirstSpace	&	Greedy	&	48.5	&	49.9	&	64.0	&	30.3	&	23.3	&	65.7	\\
Unigram	&		&	FirstSpace	&	PathPieceL	&	43.1	&	40.5	&	56.0	&	28.6	&	23.0	&	60.3	\\
\midrule

WordPiece	&		&	FirstSpace	&	Greedy	&	49.1	&	52.3	&	70.0	&	28.6	&	23.7	&	66.5	\\
\midrule

\multirow{3}{*}{SaGe}	&	BPE	&	FirstSpace	&	Greedy	&	49.2	&	50.8	&	70.0	&	29.9	&	23.2	&	66.4	\\
	&	$n$-gram	&	FirstSpDigit	&	Greedy	&	46.9	&	48.4	&	67.0	&	30.3	&	22.6	&	64.0	\\
	&	$n$-gram	&	FirstSpace	&	Greedy	&	48.5	&	49.8	&	68.0	&	\textbf{32.9}	&	22.8	&	65.4	\\
	&	Unigram	&	FirstSpace	&	Greedy	&	46.9	&	51.7	&	65.0	&	28.6	&	23.9	&	65.2	\\
\midrule

\multirow{4}{*}{PathPieceL}	&	BPE	&	FirstSpace	&	PathPieceL	&	49.4	&	52.1	&	\textbf{71.0}	&	29.9	&	23.9	&	66.9	\\
	&	$n$-gram	&	FirstSpace	&	PathPieceL	&	45.5	&	42.6	&	63.0	&	30.3	&	22.7	&	60.9	\\
	&	$n$-gram	&	FirstSpDigit	&	PathPieceL	&	44.9	&	44.0	&	60.0	&	29.9	&	22.6	&	60.8	\\
	&	Unigram	&	FirstSpace	&	PathPieceL	&	48.5	&	51.7	&	\textbf{71.0}	&	31.2	&	24.2	&	66.2	\\
\midrule

\multirow{3}{*}{PathPieceR}	&	$n$-gram	&	FirstSpDigit	&	PathPieceR	&	45.8	&	47.5	&	63.0	&	28.2	&	22.4	&	60.7	\\
	&	$n$-gram	&	None	&	PathPieceR	&	44.0	&	41.2	&	66.0	&	26.5	&	21.6	&	62.4	\\
	&	$n$-gram	&	SpaceDigit	&	PathPieceR	&	45.4	&	46.3	&	64.0	&	32.1	&	22.7	&	60.0	\\
\midrule

\multirow{8}{*}{RandTrain}	&	BPE	&	FirstSpace	&	Greedy	&	48.6	&	50.5	&	70.0	&	29.5	&	23.4	&	65.8	\\
	&	$n$-gram	&	FirstSpDigit	&	Greedy	&	47.9	&	50.0	&	63.0	&	29.5	&	23.3	&	65.3	\\
	&	$n$-gram	&	FirstSpace	&	Greedy	&	48.3	&	50.3	&	70.0	&	28.2	&	\textbf{24.3}	&	65.8	\\
	&	$n$-gram	&	None	&	Greedy	&	42.2	&	41.3	&	55.0	&	27.4	&	21.7	&	63.2	\\
	&	BPE	&	FirstSpace	&	PathPieceL	&	46.5	&	45.8	&	65.0	&	30.8	&	23.3	&	62.8	\\
	&	$n$-gram	&	FirstSpDigit	&	PathPieceL	&	38.8	&	31.2	&	48.0	&	27.8	&	22.6	&	54.7	\\
	&	$n$-gram	&	FirstSpace	&	PathPieceL	&	40.0	&	30.7	&	55.0	&	26.5	&	20.8	&	55.4	\\
	&	$n$-gram	&	None	&	PathPieceL	&	36.8	&	27.7	&	56.0	&	28.6	&	22.8	&	54.5	\\
\midrule

random	&		&		&		&	32.0	&	25.0	&	50.0	&	25.0	&	20.0	&	50.0	\\

    \bottomrule
    \end{tabular}
   \caption{350M parameter model, 40,960 token vocabulary, accuracy (\%) on average and initial 5 tasks}
    \label{tab:350_40960_pt1}
    
\end{table*}

\begin{table*}[hbt!]
    \centering
    \small
    \begin{tabular}{llllrrrrr}
    \toprule
\textbf{Vocab Constr}	&	\textbf{Init Voc}	&	\textbf{Pre-tok}	&	\textbf{Segment}	&	\textbf{qa4mre}	&	\textbf{race}	&	\textbf{sciq}	&	\textbf{sociology}	&	\textbf{wsc273}	\\
\midrule

\multirow{3}{*}{BPE}	&		&	FirstSpace	&	Merge	&	32.4	&	30.1	&	87.7	&	35.3	&	69.2	\\
	&		&	FirstSpace	&	Greedy	&	31.7	&	\textbf{30.9}	&	88.3	&	\textbf{35.8}	&	68.9	\\
	&		&	FirstSpace	&	PathPieceL	&	30.3	&	30.2	&	83.8	&	35.3	&	68.1	\\
\midrule

\multirow{3}{*}{Unigram}	&		&	FirstSpace	&	Likelihood	&	29.6	&	30.8	&	86.4	&	32.8	&	67.8	\\
	&		&	FirstSpace	&	Greedy	&	32.4	&	29.6	&	86.7	&	32.8	&	\textbf{70.3}	\\
	&		&	FirstSpace	&	PathPieceL	&	30.3	&	27.4	&	75.0	&	27.4	&	62.3	\\
\midrule

WordPiece	&		&	FirstSpace	&	Greedy	&	31.0	&	30.3	&	87.7	&	32.8	&	68.1	\\
\midrule

\multirow{4}{*}{SaGe}	&	BPE	&	FirstSpace	&	Greedy	&	28.9	&	30.2	&	\textbf{89.5}	&	34.8	&	67.8	\\
	&	$n$-gram	&	FirstSpDigit	&	Greedy	&	30.6	&	28.1	&	85.8	&	32.3	&	59.7	\\
	&	$n$-gram	&	FirstSpace	&	Greedy	&	29.2	&	30.0	&	88.4	&	33.3	&	65.2	\\
	&	Unigram	&	FirstSpace	&	Greedy	&	26.8	&	29.1	&	86.9	&	31.3	&	60.1	\\
\midrule

\multirow{4}{*}{PathPieceL}	&	BPE	&	FirstSpace	&	PathPieceL	&	31.0	&	29.6	&	87.3	&	34.3	&	67.8	\\
	&	$n$-gram	&	FirstSpace	&	PathPieceL	&	29.9	&	27.9	&	81.0	&	34.8	&	61.9	\\
	&	$n$-gram	&	FirstSpDigit	&	PathPieceL	&	27.5	&	28.2	&	80.7	&	30.9	&	64.1	\\
	&	Unigram	&	FirstSpace	&	PathPieceL	&	31.3	&	29.7	&	86.3	&	29.9	&	63.7	\\
\midrule

\multirow{3}{*}{PathPieceR}	&	$n$-gram	&	FirstSpDigit	&	PathPieceR	&	29.9	&	30.8	&	82.1	&	27.4	&	66.3	\\
	&	$n$-gram	&	None	&	PathPieceR	&	23.6	&	28.3	&	73.8	&	\textbf{35.8}	&	60.4	\\
	&	$n$-gram	&	SpaceDigit	&	PathPieceR	&	27.5	&	28.7	&	78.2	&	31.3	&	63.0	\\
\midrule

\multirow{8}{*}{RandTrain}	&	BPE	&	FirstSpace	&	Greedy	&	32.0	&	29.6	&	86.9	&	30.9	&	67.4	\\
	&	$n$-gram	&	FirstSpDigit	&	Greedy	&	30.6	&	30.0	&	87.5	&	31.3	&	68.1	\\
	&	$n$-gram	&	FirstSpace	&	Greedy	&	29.9	&	29.7	&	85.3	&	32.8	&	67.0	\\
	&	$n$-gram	&	None	&	Greedy	&	28.2	&	27.8	&	75.9	&	26.4	&	55.0	\\
	&	BPE	&	FirstSpace	&	PathPieceL	&	\textbf{32.8}	&	28.5	&	80.3	&	30.9	&	64.5	\\
	&	$n$-gram	&	FirstSpDigit	&	PathPieceL	&	31.3	&	24.2	&	62.1	&	30.4	&	55.3	\\
	&	$n$-gram	&	FirstSpace	&	PathPieceL	&	28.9	&	23.6	&	66.8	&	33.8	&	59.0	\\
	&	$n$-gram	&	None	&	PathPieceL	&	21.5	&	24.9	&	51.8	&	28.9	&	51.7	\\
\midrule

random	&		&		&		&	25.0	&	25.0	&	25.0	&	25.0	&	50.0	\\

    \bottomrule
    \end{tabular}
   \caption{350M parameter model, 40,960 token vocabulary, accuracy (\%) on remaining 5 tasks}
    \label{tab:350_40960_pt2}
    
\end{table*}

\begin{table*}[hbt!]
    \centering
    \small
    \begin{tabular}{llllrrrrrr}
    \toprule
\textbf{Vocab Constr}	&	\textbf{Init Voc}	&	\textbf{Pre-tok}	&	\textbf{Segment}	&	\textbf{Avg}	&	\textbf{arc\_easy}	&	\textbf{copa}	&	\textbf{mktg}	&	\textbf{mathqa}	&	\textbf{piqa}	\\
\midrule

\multirow{3}{*}{BPE}	&		&	FirstSpace	&	Merge	&	48.1	&	52.3	&	65.0	&	31.6	&	23.7	&	65.7	\\
	&		&	FirstSpace	&	Greedy	&	\textbf{49.5}	&	\textbf{53.9}	&	\textbf{72.0}	&	31.6	&	24.2	&	\textbf{68.4}	\\
	&		&	FirstSpace	&	PathPieceL	&	47.2	&	48.6	&	69.0	&	26.9	&	22.8	&	63.1\\
\midrule

\multirow{3}{*}{Unigram}	&		&	FirstSpace	&	Likelihood	&	48.8	&	52.3	&	69.0	&	\textbf{35.0}	&	23.9	&	66.1	\\
	&		&	FirstSpace	&	Greedy	&	48.6	&	51.6	&	68.0	&	32.1	&	24.4	&	65.7	\\
	&		&	FirstSpace	&	PathPieceL	&	44.0	&	39.4	&	57.0	&	30.3	&	23.3	&	61.2	\\
\midrule

WordPiece	&		&	FirstSpace	&	Greedy	&	48.8	&	52.6	&	68.0	&	28.2	&	23.5	&	66.2	\\
\midrule

\multirow{4}{*}{SaGe}	&	BPE	&	FirstSpace	&	Greedy	&	48.8	&	51.9	&	71.0	&	29.9	&	22.6	&	65.5	\\
	&	$n$-gram	&	FirstSpDigit	&	Greedy	&	47.2	&	46.6	&	67.0	&	31.2	&	22.7	&	63.4	\\
	&	$n$-gram	&	FirstSpace	&	Greedy	&	48.0	&	49.7	&	66.0	&	31.6	&	21.6	&	65.7	\\
	&	Unigram	&	FirstSpace	&	Greedy	&	47.8	&	49.7	&	68.0	&	29.9	&	23.5	&	64.6	\\
\midrule

\multirow{4}{*}{PathPieceL}	&	BPE	&	FirstSpace	&	PathPieceL	&	49.4	&	51.9	&	69.0	&	29.9	&	24.5	&	66.6	\\
	&	$n$-gram	&	FirstSpace	&	PathPieceL	&	43.9	&	42.4	&	56.0	&	28.6	&	23.8	&	60.3	\\
	&	$n$-gram	&	FirstSpDigit	&	PathPieceL	&	45.0	&	44.5	&	59.0	&	28.2	&	22.3	&	59.5	\\
	&	Unigram	&	FirstSpace	&	PathPieceL	&	48.4	&	51.4	&	67.0	&	29.5	&	\textbf{24.7}	&	65.2	\\
\midrule

\multirow{3}{*}{PathPieceR}	&	$n$-gram	&	FirstSpDigit	&	PathPieceR	&	45.5	&	46.0	&	62.0	&	25.6	&	22.1	&	61.6	\\
	&	$n$-gram	&	None	&	PathPieceR	&	42.2	&	42.6	&	64.0	&	22.2	&	22.4	&	60.9	\\
	&	$n$-gram	&	SpaceDigit	&	PathPieceR	&	47.3	&	48.7	&	68.0	&	34.2	&	21.9	&	65.1	\\
\midrule
random	&		&		&		&	32.0	&	25.0	&	50.0	&	25.0	&	20.0	&	50.0	\\

    \bottomrule
    \end{tabular}
   \caption{350M parameter model, 49,152 token vocabulary, accuracy (\%) on average and initial 5 tasks}
    \label{tab:350_49152_pt1}
    
\end{table*}

\begin{table*}[hbt!]
    \centering
    \small
    \begin{tabular}{llllrrrrr}
    \toprule
\textbf{Vocab Constr}	&	\textbf{Init Voc}	&	\textbf{Pre-tok}	&	\textbf{Segment}	&	\textbf{qa4mre}	&	\textbf{race}	&	\textbf{sciq}	&	\textbf{sociology}	&	\textbf{wsc273}	\\
\midrule

\multirow{3}{*}{BPE}	&		&	FirstSpace	&	Merge	&	28.9	&	31.0	&	87.3	&	28.9	&	67.0	\\
	&		&	FirstSpace	&	Greedy	&	29.6	&	31.2	&	88.4	&	29.4	&	66.3	\\
	&		&	FirstSpace	&	PathPieceL	&	31.0	&	30.7	&	85.4	&	31.8	&	63.0	\\
\midrule

\multirow{3}{*}{Unigram}	&		&	FirstSpace	&	Likelihood	&	27.5	&	30.3	&	\textbf{89.1}	&	28.9	&	65.9	\\
	&		&	FirstSpace	&	Greedy	&	32.4	&	29.5	&	86.7	&	32.3	&	63.7	\\
	&		&	FirstSpace	&	PathPieceL	&	\textbf{33.1}	&	26.0	&	74.5	&	27.9	&	67.0	\\
\midrule

WordPiece	&		&	FirstSpace	&	Greedy	&	29.2	&	31.1	&	88.0	&	34.3	&	66.7	\\
\midrule

\multirow{4}{*}{SaGe}	&	BPE	&	FirstSpace	&	Greedy	&	29.6	&	31.2	&	87.5	&	32.3	&	65.9	\\
	&	$n$-gram	&	FirstSpDigit	&	Greedy	&	29.2	&	28.8	&	86.4	&	34.3	&	61.9	\\
	&	$n$-gram	&	FirstSpace	&	Greedy	&	28.8	&	30.2	&	87.5	&	33.8	&	64.5	\\
	&	Unigram	&	FirstSpace	&	Greedy	&	28.9	&	\textbf{31.4}	&	87.0	&	29.9	&	65.6	\\
\midrule

\multirow{4}{*}{PathPieceL}	&	BPE	&	FirstSpace	&	PathPieceL	&	31.0	&	\textbf{31.4}	&	87.5	&	31.3	&	\textbf{70.7}	\\
	&	$n$-gram	&	FirstSpace	&	PathPieceL	&	27.5	&	26.7	&	80.8	&	32.3	&	60.8	\\
	&	$n$-gram	&	FirstSpDigit	&	PathPieceL	&	28.9	&	30.0	&	80.6	&	\textbf{35.8}	&	61.2	\\
	&	Unigram	&	FirstSpace	&	PathPieceL	&	29.2	&	30.5	&	88.5	&	32.8	&	65.6	\\
\midrule

\multirow{3}{*}{PathPieceR}	&	$n$-gram	&	FirstSpDigit	&	PathPieceR	&	29.6	&	29.5	&	82.8	&	30.9	&	64.5	\\
	&	$n$-gram	&	None	&	PathPieceR	&	25.7	&	27.5	&	72.5	&	27.4	&	57.1	\\
	&	$n$-gram	&	SpaceDigit	&	PathPieceR	&	27.5	&	28.7	&	84.0	&	28.9	&	66.3	\\
\midrule

Random	&		&		&		&	25.0	&	25.0	&	25.0	&	25.0	&	50.0	\\

    \bottomrule
    \end{tabular}
   \caption{350M parameter model, 49,152 token vocabulary, accuracy (\%) on remaining 5 tasks}
    \label{tab:350_49152_pt2}
    
\end{table*}

\begin{table*}[!htb]
    \centering
    \small
    \begin{tabular}{rllllrrrr}
    \toprule
\textbf{Rank}	&	\textbf{Vocab Constr}	&	\textbf{Init Voc}	&	\textbf{Pre-tok}	&	\textbf{Segment}	&	\textbf{Overall avg}	&	\textbf{32,768 avg}	&	\textbf{40,960 avg}	&	\textbf{49,152 avg}	\\

\midrule

1	&	PathPieceL	&	BPE	&	FirstSpace	&	PathPieceL	&	\textbf{49.4}	&	\textbf{49.3}	&	49.4	&	49.4	\\
2	&	Unigram	&		&	FirstSpace	&	Likelihood	&	49.0	&	49.2	&	49.1	&	48.8	\\
3	&	BPE	&		&	FirstSpace	&	Merge	&	49.0	&	48.8	&	\textbf{50.0}	&	48.1	\\
4	&	BPE	&		&	FirstSpace	&	Greedy	&	49.0	&	48.3	&	49.1	&	\textbf{49.5}	\\
5	&	WordPiece	&		&	FirstSpace	&	Greedy	&	48.8	&	48.5	&	49.1	&	48.8	\\
6	&	SaGe	&	BPE	&	FirstSpace	&	Greedy	&	48.6	&	47.9	&	49.2	&	48.8	\\
7	&	Unigram	&		&	FirstSpace	&	Greedy	&	48.3	&	47.9	&	48.5	&	48.6	\\
8	&	SaGe	&	$n$-gram	&	FirstSpace	&	Greedy	&	48.0	&	47.5	&	48.5	&	48.0	\\
9	&	PathPieceL	&	Unigram	&	FirstSpace	&	PathPieceL	&	48.0	&	46.9	&	48.5	&	48.4	\\
10	&	SaGe	&	Unigram	&	FirstSpace	&	Greedy	&	47.7	&	48.4	&	46.9	&	47.8	\\
11	&	SaGe	&	$n$-gram	&	FirstSpDigit	&	Greedy	&	47.5	&	48.4	&	46.9	&	47.2	\\
12	&	PathPieceR	&	$n$-gram	&	SpaceDigit	&	PathPieceR	&	46.7	&	47.5	&	45.4	&	47.3	\\
13	&	BPE	&		&	FirstSpace	&	PathPieceL	&	46.5	&	45.6	&	46.7	&	47.2	\\
14	&	PathPieceR	&	$n$-gram	&	FirstSpDigit	&	PathPieceR	&	45.5	&	45.3	&	45.8	&	45.5	\\
15	&	PathPieceL	&	$n$-gram	&	FirstSpDigit	&	PathPieceL	&	44.8	&	44.6	&	44.9	&	45.0	\\
16	&	PathPieceL	&	$n$-gram	&	FirstSpace	&	PathPieceL	&	44.7	&	44.8	&	45.5	&	43.9	\\
17	&	Unigram	&		&	FirstSpace	&	PathPieceL	&	43.6	&	43.6	&	43.1	&	44.0	\\
18	&	PathPieceR	&	$n$-gram	&	None	&	PathPieceR	&	43.2	&	43.5	&	44.0	&	42.2	\\
\midrule
		&	Random	&		&		&		&	32.0 & 32.0 & 32.0 & 32.0	\\

    \bottomrule
    \end{tabular}
   \caption{Summary of 350M parameter model downstream accuracy (\%), sorted by rank}
    \label{tab:overallavg}
    
\end{table*}

\begin{table*}[!htb]
    \centering
    \small
    \begin{tabular}{rrrrrrrrr}
    \toprule
\textbf{Rank} & \textbf{Vocab Size} & \textbf{Avg Acc} & \textbf{CTC} & \textbf{Eff $\alpha$=1.5} & \textbf{Eff $\alpha$=2} & \textbf{Eff $\alpha$=2.5} & \textbf{Eff $\alpha$=3} & \textbf{Eff $\alpha$=3.5} \\
\midrule
1 &  32,768  & 49.3 & 1.48 & 0.604 & 0.516 & 0.469 & 0.441 & 0.422 \\
1 &  40,960  & 49.4 & 1.46 & 0.589 & 0.503 & 0.457 & 0.429 & 0.411 \\
1 &  49,152  & 49.4 & 1.44 & 0.578 & 0.492 & 0.448 & 0.420 & 0.402 \\
2 &  32,768  & 49.2 & 1.79 & 0.461 & 0.371 & 0.324 & 0.295 & 0.277 \\
2 &  40,960  & 49.1 & 1.77 & 0.451 & 0.362 & 0.316 & 0.289 & 0.271 \\
2 &  49,152  & 48.8 & 1.76 & 0.444 & 0.356 & 0.311 & 0.284 & 0.266 \\
3 &  32,768  & 48.8 & 1.52 & 0.594 & 0.505 & 0.459 & 0.431 & 0.414 \\
3 &  40,960  & 50.0 & 1.49 & 0.579 & 0.491 & 0.446 & 0.420 & 0.403 \\
3 &  49,152  & 48.1 & 1.47 & 0.567 & 0.481 & 0.437 & 0.411 & 0.394 \\
4 &  32,768  & 48.3 & 1.50 & 0.605 & 0.517 & 0.471 & 0.442 & 0.423 \\
4 &  40,960  & 49.1 & 1.48 & 0.590 & 0.504 & 0.458 & 0.430 & 0.412 \\
4 &  49,152  & 49.5 & 1.46 & 0.579 & 0.494 & 0.449 & 0.421 & 0.403 \\
5 &  32,768  & 48.5 & 1.54 & 0.598 & 0.507 & 0.461 & 0.433 & 0.415 \\
5 &  40,960  & 49.1 & 1.51 & 0.583 & 0.494 & 0.448 & 0.421 & 0.404 \\
5 &  49,152  & 48.8 & 1.49 & 0.571 & 0.483 & 0.439 & 0.412 & 0.396 \\
6 &  32,768  & 47.9 & 1.78 & 0.545 & 0.466 & 0.422 & 0.396 & 0.378 \\
6 &  40,960  & 49.2 & 1.76 & 0.533 & 0.455 & 0.413 & 0.387 & 0.369 \\
6 &  49,152  & 48.7 & 1.75 & 0.523 & 0.447 & 0.405 & 0.379 & 0.362 \\
7 &  32,768  & 47.9 & 1.81 & 0.510 & 0.431 & 0.387 & 0.359 & 0.340 \\
7 &  40,960  & 48.5 & 1.79 & 0.500 & 0.423 & 0.381 & 0.354 & 0.335 \\
7 &  49,152  & 48.6 & 1.77 & 0.493 & 0.416 & 0.375 & 0.348 & 0.330 \\
8 &  32,768  & 47.5 & 1.63 & 0.629 & 0.536 & 0.482 & 0.447 & 0.424 \\
8 &  40,960  & 48.5 & 1.62 & 0.615 & 0.524 & 0.470 & 0.437 & 0.415 \\
8 &  49,152  & 48.0 & 1.62 & 0.605 & 0.515 & 0.462 & 0.429 & 0.407 \\
9 &  32,768  & 46.9 & 1.74 & 0.508 & 0.419 & 0.372 & 0.343 & 0.323 \\
9 &  40,960  & 48.5 & 1.72 & 0.491 & 0.403 & 0.356 & 0.328 & 0.309 \\
9 &  49,152  & 48.4 & 1.72 & 0.477 & 0.389 & 0.343 & 0.315 & 0.296 \\
10 &  32,768  & 48.4 & 2.02 & 0.485 & 0.409 & 0.366 & 0.339 & 0.320 \\
10 &  40,960  & 46.9 & 2.01 & 0.474 & 0.401 & 0.358 & 0.331 & 0.313 \\
10 &  49,152  & 47.8 & 2.01 & 0.466 & 0.393 & 0.352 & 0.325 & 0.307 \\
11 &  32,768  & 48.4 & 1.77 & 0.587 & 0.512 & 0.470 & 0.443 & 0.425 \\
11 &  40,960  & 46.9 & 1.76 & 0.575 & 0.501 & 0.460 & 0.433 & 0.415 \\
11 &  49,152  & 47.2 & 1.76 & 0.565 & 0.492 & 0.452 & 0.426 & 0.408 \\
12 &  32,768  & 47.5 & 2.33 & 0.236 & 0.164 & 0.138 & 0.124 & 0.116 \\
12 &  40,960  & 45.4 & 2.30 & 0.228 & 0.159 & 0.133 & 0.120 & 0.112 \\
12 &  49,152  & 47.3 & 2.29 & 0.223 & 0.155 & 0.130 & 0.117 & 0.109 \\
13 &  32,768  & 45.6 & 1.50 & 0.606 & 0.518 & 0.470 & 0.442 & 0.423 \\
13 &  40,960  & 46.7 & 1.47 & 0.591 & 0.504 & 0.458 & 0.430 & 0.412 \\
13 &  49,152  & 47.2 & 1.45 & 0.579 & 0.494 & 0.449 & 0.421 & 0.403 \\
14 &  32,768  & 45.3 & 1.46 & 0.616 & 0.532 & 0.490 & 0.465 & 0.448 \\
14 &  40,960  & 45.8 & 1.43 & 0.602 & 0.519 & 0.478 & 0.453 & 0.437 \\
14 &  49,152  & 45.5 & 1.42 & 0.591 & 0.508 & 0.468 & 0.444 & 0.428 \\
15 &  32,768  & 44.6 & 1.47 & 0.620 & 0.533 & 0.490 & 0.464 & 0.447 \\
15 &  40,960  & 44.9 & 1.44 & 0.605 & 0.520 & 0.478 & 0.453 & 0.436 \\
15 &  49,152  & 45.0 & 1.42 & 0.594 & 0.509 & 0.468 & 0.443 & 0.427 \\
16 &  32,768  & 44.8 & 1.36 & 0.677 & 0.571 & 0.514 & 0.480 & 0.457 \\
16 &  40,960  & 45.5 & 1.33 & 0.662 & 0.556 & 0.500 & 0.466 & 0.444 \\
16 &  49,152  & 43.9 & 1.31 & 0.650 & 0.544 & 0.489 & 0.456 & 0.435 \\
17 &  32,768  & 43.6 & 1.77 & 0.471 & 0.380 & 0.333 & 0.304 & 0.285 \\
17 &  40,960  & 43.1 & 1.75 & 0.462 & 0.372 & 0.326 & 0.298 & 0.280 \\
17 &  49,152  & 44.0 & 1.74 & 0.455 & 0.366 & 0.320 & 0.293 & 0.275 \\
18 &  32,768  & 43.5 & 1.29 & 0.747 & 0.617 & 0.549 & 0.511 & 0.486 \\
18 &  40,960  & 44.0 & 1.26 & 0.736 & 0.603 & 0.535 & 0.497 & 0.474 \\
18 &  49,152  & 42.2 & 1.25 & 0.728 & 0.591 & 0.524 & 0.487 & 0.464 \\

   \bottomrule
    \end{tabular}
   \caption{Average Accuracy (\%) vs. Corpus Token Count (CTC, in billions) by vocabulary size, for \autoref{fig:total_tok_vs_accuracy}. 
   Also includes the corresponding R\'enyi efficiency \citep{zouhar-etal-2023-tokenization} for various orders $\alpha$.}
    \label{tab:fig3data}
    
\end{table*}

\begin{table*}[hbt!]
    \centering
    \small
    \begin{tabular}{llllrrrrrr}
    \toprule
\textbf{Vocab Constr}	&	\textbf{Init Voc}	&	\textbf{Pre-tok}	&	\textbf{Segment}	&	\textbf{Avg}	&	\textbf{arc\_easy}	&	\textbf{copa}	&	\textbf{mktg}	&	\textbf{mathqa}	&	\textbf{piqa}	\\
\midrule

BPE	&		&	FirstSpace	&	Merge	&	\textbf{53.1}	&	\textbf{62.0}	&	\textbf{77.0}	&	\textbf{32.1}	&	25.0	&	71.1	\\
\midrule

Unigram	&		&	FirstSpace	&	Likelihood	&	52.4	&	60.6	&	71.0	&	30.3	&	\textbf{25.2}	&	71.0	\\
\midrule

\multirow{2}{*}{SaGe}	&	BPE	&	FirstSpace	&	Greedy	&	52.2	&	62.0	&	72.0	&	27.4	&	24.5	&	\textbf{71.6}	\\
	&	$n$-gram	&	FirstSpDigit	&	Greedy	&	50.7	&	60.3	&	71.0	&	28.6	&	22.8	&	69.4	\\
\midrule

\multirow{3}{*}{PathPieceL}	&	BPE	&	FirstSpace	&	PathPieceL	&	49.2	&	57.4	&	66.0	&	27.8	&	24.3	&	65.9	\\
	&	$n$-gram	&	FirstSpDigit	&	PathPieceL	&	47.6	&	49.7	&	67.0	&	24.8	&	23.4	&	63.2	\\
	&	$n$-gram	&	SpaceDigit	&	PathPieceL	&	46.3	&	51.1	&	59.0	&	28.6	&	23.3	&	63.8	\\
\midrule

Random	&		&		&		&	32.0	&	25.0	&	50.0	&	25.0	&	20.0	&	50.0	\\

    \bottomrule
    \end{tabular}
   \caption{1.3B parameter model, 40,960 token vocabulary, accuracy (\%) on average and initial 5 tasks}
    \label{tab:1.2_pt1}
    
\end{table*}

\begin{table*}[hbt!]
    \centering
    \small
    \begin{tabular}{llllrrrrr}
    \toprule
\textbf{Vocab Constr}	&	\textbf{Init Voc}	&	\textbf{Pre-tok}	&	\textbf{Segment}	&	\textbf{qa4mre}	&	\textbf{race}	&	\textbf{sciq}	&	\textbf{sociology}	&	\textbf{wsc273}	\\
\midrule

BPE	&		&	FirstSpace	&	Merge	&	32.4	&	\textbf{34.9}	&	\textbf{93.0}	&	26.4	&	\textbf{76.9}	\\
\midrule

Unigram	&		&	FirstSpace	&	Likelihood	&	\textbf{37.7}	&	33.0	&	91.8	&	28.9	&	74.4	\\
\midrule

\multirow{2}{*}{SaGe}	&	BPE	&	FirstSpace	&	Greedy	&	34.9	&	34.8	&	92.5	&	25.9	&	76.2	\\
	&	$n$-gram	&	FirstSpDigit	&	Greedy	&	29.9	&	32.9	&	91.5	&	\textbf{29.4}	&	71.1	\\
\midrule

\multirow{3}{*}{PathPieceL}	&	BPE	&	FirstSpace	&	PathPieceL	&	31.0	&	33.3	&	89.4	&	26.4	&	70.7	\\
	&	$n$-gram	&	FirstSpDigit	&	PathPieceL	&	31.0	&	31.6	&	86.1	&	\textbf{29.4}	&	70.0	\\
	&	$n$-gram	&	SpaceDigit	&	PathPieceL	&	28.9	&	31.3	&	87.1	&	22.4	&	67.0	\\
\midrule

Random	&		&		&		&	25.0	&	25.0	&	25.0	&	25.0	&	50.0	\\

    \bottomrule
    \end{tabular}
   \caption{1.3B parameter model, 40,960 token vocabulary, accuracy (\%) on remaining 5 tasks}
    \label{tab:1.2_pt2}
    
\end{table*}

\begin{table*}[!ht]
    \centering
    \small
    \begin{tabular}{llllrrrrrr}
    \toprule
\textbf{Voc Con}	&	\textbf{Init V}	&	\textbf{Pre-tok}	&	\textbf{Seg}	&	\textbf{350M avg} & \textbf{350M rnk} & \textbf{1.3B avg} & \textbf{1.3B rnk}	& \textbf{2.4B avg}  & \textbf{2.4B rnk} \\

\midrule

BPE	    &		        &	FirSp	&	Merge	& 50.0	& 1 & 53.1 & 1  & 54.2 & 3	\\
\midrule
PathPL	&	BPE	        &	FirSp	&	PathPL	& 49.4	& 3 & 49.2 & 5  & 52.7 & 4  \\
PathPL	&	$n$-gram	&	FirSpD	&	PathPL	& 44.9	& 6 & 47.6 &	6  \\
\midrule
SaGe	&	BPE	        &	FirSp	&	Greedy	& 49.2	& 2 & 52.2 & 3  &  55.0 & 1	\\
SaGe	&	$n$-gram	&	FirSpD	&	Greedy	& 46.9 & 5 & 50.7 & 4	\\
\midrule
Unigram	&		        &	FirSp	&	Likeli	& 49.1 & 4 & 52.4 & 2& 54.7 & 2      \\

    \bottomrule
    \end{tabular}
   \caption{Downstream accuracy (\%) of 10 tasks with vocab size 40,960, for various model sizes}
    \label{tab:summary1.2}
    
\end{table*}

% this doesn't line up with anything else, even a 350M picked kind of random one 
% PathPL	&	$n$-gram	&	SpD	    &	PathPL	&	&  & 46.3 & 	7  \\  

\end{document}